\documentclass{article}

\usepackage[final]{neurips_2022}

\usepackage[utf8]{inputenc} 
\usepackage[T1]{fontenc}    
\usepackage{hyperref}       
\usepackage{url}            
\usepackage{booktabs}       
\usepackage{amsfonts}       
\usepackage{nicefrac}       
\usepackage{microtype}      
\usepackage{xcolor}         
\usepackage{multirow}

\UseRawInputEncoding
\usepackage{tikz-cd}
\usepackage{verbatim}
\usepackage{amsmath}
\usepackage{amssymb}
\usepackage{mathtools}
\usepackage{amsthm}
\usepackage{wrapfig}

\title{Unsupervised Learning of Group Invariant and Equivariant Representations}

\author{%
  Robin Winter\thanks{equal contribution}\\
  Bayer AG \\
  Freie Universit\"{a}t Berlin \\
  \texttt{robin.winter@bayer.com}\\
  \And
  Marco Bertolini$^*$\\
  Bayer AG \\
  \texttt{marco.bertolini@bayer.com}\\
  \And
  Tuan Le\\
  Bayer AG \\
  Freie Universit\"{a}t Berlin \\
  \texttt{tuan.le2@bayer.com}\\
  \And
  Frank No\'{e} \\
  Freie Universit\"{a}t Berlin\\
  Microsoft Research \\
  \texttt{frank.noe@fu-berlin.de}
  \And
  Djork-Arn\'{e} Clevert \\
  Bayer AG\\
  \texttt{djork-arne.clevert@bayer.com} \\
}

\theoremstyle{plain}
\newtheorem{theorem}{Theorem}[section]
\newtheorem{proposition}[theorem]{Proposition}
\newtheorem{property}[theorem]{Property}
\newtheorem{lemma}[theorem]{Lemma}

\theoremstyle{definition}

\theoremstyle{remark}

\newcommand{\SO}{\text{SO}}
\newcommand{\SE}{\text{SE}}
\newcommand{\GL}{\text{GL}}
\newcommand{\R}{\mathbb{R}}
\newcommand{\Z}{\mathbb{Z}}

\newcommand{\Imag}{\text{Im}}

\newcommand{\angstrom}{\mbox{\normalfont\AA}}

\begin{document}

\maketitle

\begin{abstract}
Equivariant neural networks, whose hidden features transform according to representations of a group $G$ acting on the data, exhibit training efficiency and an improved generalisation performance.
In this work, we extend group invariant and equivariant representation learning to the field of unsupervised deep learning. 
We propose a general learning strategy based on an encoder-decoder framework in which the latent representation
is separated in an invariant term and an equivariant group action component. 
The key idea is that the network learns to encode and decode data to and
from a group-invariant representation by additionally learning to predict the appropriate group action to align input and output pose to solve the reconstruction
task. 
We derive the necessary conditions on the equivariant encoder, and we present a
construction valid for any $G$, both discrete and continuous. We describe explicitly our construction for rotations, translations and permutations. We test the validity and the robustness of our approach in a variety of experiments with diverse data types employing different network architectures. 
\end{abstract}

\section{Introduction}


An increasing body of work has shown that incorporating knowledge about underlying symmetries in neural networks as inductive bias can drastically improve the performance and reduce the amount of data needed for training \cite{cohen2016group, bronstein2021geometric}. For example, the equivariant design with respect to the translation symmetry of objects in images proper of convolutional neural networks (CNNs) has revolutionized the field of image analysis \cite{lecun1995convolutional}. Message Passing neural networks, respecting permutation symmetries in graphs, have enabled powerful predictive models on graph-structured data  \cite{gilmer2017neural, defferrard2016convolutional}. Recently, much work has been done utilizing 3D rotation and translation equivariant neural networks for point clouds and volumetric data, showing great success in predicting molecular ground state energy levels with high fidelity \cite{miller2020relevance, anderson2019cormorant, klicpera2020directional, schuett2021equivariant}. Invariant models take advantage of the fact that often properties of interest, such as the class label of an object in an image or the ground state energy of a molecule, are invariant to certain group actions (e.g., translations or rotations), while the data itself is not (e.g., pixel values, atom coordinates).

\vspace{-0.1cm}
There are several approaches to incorporate invariance into the learned representation of a neural network. The most common approach consists of teaching invariance to the model by data augmentation: during training, the model
must learn that a group transformation on its input does not affect its label. While this approach can lead to improved generalization performance, it reduces training efficiency and quickly becomes impractical for higher dimensional data \cite{thomas2018tensor}.
A second technique, known as feature averaging, consists of averaging model predictions over group transformations of the input \cite{Omriframeaveraging}. While feasible with finite groups, this method requires, for instance, sampling for infinite groups \cite{Lyle2020OnTB}. 
A third approach is to impose invariance as a model architectural design. The simplest option is 
to restrict the function to be learned to be a composition of symmetric functions only \cite{schutt2018schnet}.
Such choice, however, can significantly restrict the functional form of the network. A more expressive variation of this approach consists of an equivariant neural network, followed by a symmetric function. 
This allows the network to leverage the benefits of invariance while having a larger capacity due to the less restrictive nature of equivariance.
In fact, in many real-world application, equivariance is beneficial if not necessary \cite{smidt_2020, miller2020relevance}. For example, the interaction of a molecule (per se rotational invariant) with an external magnetic field is an intrinsically equivariant problem. 


\begin{figure}[t]
     \centering
         \includegraphics[width=0.99\textwidth]{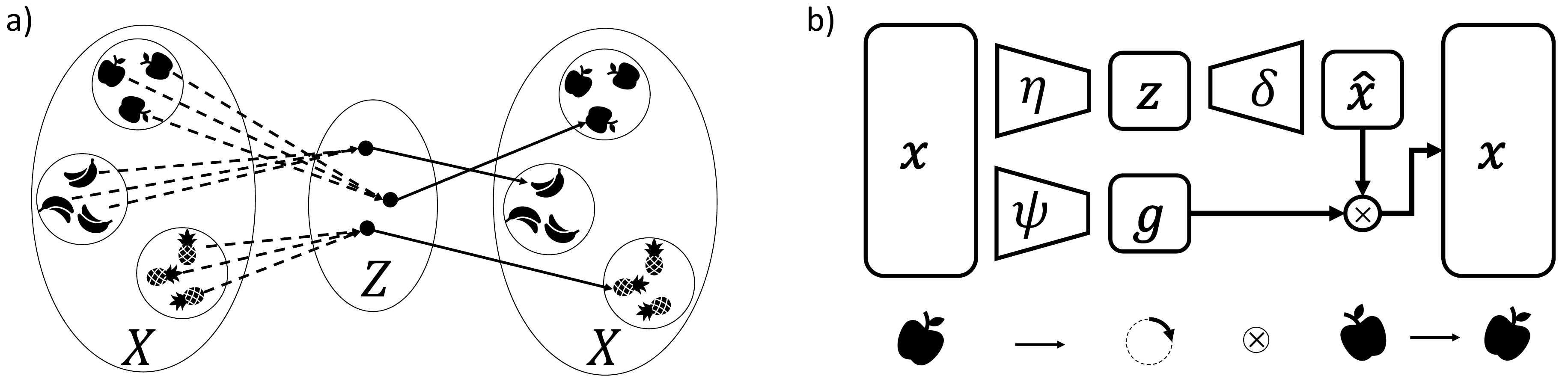}
         \caption{a) Schematic of the learning task this work is concerned with. Data points $x\in X$ are encoded to and decoded from latent space $Z$. Points in the same orbit in $X$ are mapped to the same point (orbit) $z\in Z = X/G$. Latent points $z$ are mapped to canonical elements $\hat{x}\in \{\rho_X(g)x|\forall g \in G\}$. b) Schematic of our proposed framework with data points $x$, encoding function $\eta$, decoding function $\delta$, canonical elements $\hat{x}$, group function $\psi$ and group action $g$}
         \label{fig:main}
\end{figure}

All aforementioned considerations require some sort of supervision to extract invariant representations from data. Unsupervised learning of group invariant representations, despite its potential in the field of representation learning, has been impaired by the fact that the representation of the data in general does not manifestly exhibit the group as a symmetry. 
For instance, in the case of an encoder-decoder framework in which the bottleneck layer is invariant, the reconstruction is only possible up to a group
transformation. Nevertheless, the input data is typically parametrized in terms of coordinates in some vector space $X$, and the reconstruction task can only succeed by employing knowledge about the group action on $X$.

Following this line of thought, this work is concerned with the question: \emph{Can we learn to extract both the invariant and 
the (complementary) equivariant representations of data in an unsupervised way?}

To this end, we introduce a group-invariant representation learning method that encodes data in a group-invariant latent code and a group action. By separating the embedding in a group-invariant and a group-equivariant part, we can learn expressive lower-dimensional group-invariant representations utilizing the power of autoencoders (AEs).
We can summarize the main contributions of this work as follows:
\begin{itemize}
\item We introduce a novel framework for learning group equivariant representations. Our representations are \textit{by construction} separated in an
invariant and equivariant component.

\item We characterize the mathematical conditions of the group action function component and 
we propose an explicit construction suitable for \textit{any} group $G$. To the best of our knowledge, this is the first method for unsupervised learning of separated invariant-equivariant representations valid for any group.
\item We show in various experiments the validity and flexibility of our framework by learning
representations of diverse data types with different network architectures. We also show that the invariant representations
are superiour to the non-invariant counterparts in downstream tasks, and that they can be successfully employed in transfer learning
for molecular property predictions.   
\end{itemize}

\section{Method}

\subsection{Background}

We begin this section by introducing the basic concepts which will be central in our work. 

A group $G$ is a set equipped with an operation (here denoted $\cdot$) which is associative as well as having an
identity element $e$ and inverse elements.
In the context of data, we are mainly interested in how groups represent geometric transformations by acting on spaces and, in particular, 
how they describe the symmetries of an object or of a set. 
In either case, we are interested in how groups act on spaces. This
is represented by a \textbf{group action}: given a set $X$ and a group $G$, a (left) action of $G$ on $X$ is a map 
$\rho: G\times X \rightarrow X$ such that it respects the group property of associativity and identity element.
If $X$ is a vector
space, which we will assume for the remainder of the text, 
we refer to group actions of the form $\rho_X: G \rightarrow \GL(X)$ as \textbf{representations} of $G$, where the general linear group of degree $n$ $\GL(X)$ is represented by the set of $n\times n$ invertible matrices. Given a group action, a concept which will play an important role in our discussion is given by the fixed points of such an action. Formally, given a point $x\in X$ and an action (representation) $\rho_X$ of $G$ on $X$, the \textbf{stabilizer} of $G$ with respect to $x$ is the subgroup $G_x=\{g\in G | \rho_X(g)x = x\}\subset G$.

In the context of representation learning, 
we assume our data to be defined 
as the space of representation-valued functions on some set $V$, i.e., $X = \{ f | f: V \rightarrow W\}$. 
For instance, a point cloud in three dimensions can be represented as 
the set of functions $f:\R^3 \rightarrow \Z_2$, assigning to every point $\mathbf{r}\in \R^3$ the value $f(\mathbf{r})=0$ (the point is not
included in the cloud) or $f(\mathbf{r})=1$ (the point is included in the cloud). 
Representations $\rho_V$ of a group $G$ on $V$ can be extended to representations on $f$, and therefore on $X$,  $\rho_X: G \rightarrow \GL(X)$, as follows
\begin{align}
\label{eq:reprcomps}
 \left[\rho_X(g)f\right](x) \equiv \rho_W(g) f(\rho_V(g^{-1})x)~.
\end{align}
In what follows, we will then only refer to representations for the space $X$, implicitly referring to equation \eqref{eq:reprcomps} for mapping back to how the various components transform.
A map $\varphi: V \rightarrow W$ is said to be \textbf{$G$-equivariant} with respect to the actions (representations) $\rho_V, \rho_W$
if $\varphi(\rho_V(g)v) = \rho_W(g) \varphi(v)$ for every $g\in G$ and $v \in V$.
Note that $G$-invariance is a particular case of the above, where we take $\rho_V,\rho_W$ to be the trivial representations. 
An element $x\in X$ can be described in terms of a $G$-invariant component and a group element $g\in G$, as follows: let 
$\varphi_{\text{inv}}: X \rightarrow X/G$, be an invariant map mapping each element $x\in X$ to a corresponding 
canonical element $\hat{x}$ in the orbit in the quotient space $X/G$. 
Then for each $x\in X$ there exist a $g\in G$ such that $x = \rho_X(g)\varphi_{\text{inv}}(x)$. 

\subsection{Problem Definition}
\label{s:problem}
We consider a classical autoencoder framework with encoding function $\eta:X\rightarrow Z$ and decoding function $\delta:Z\rightarrow X$, mapping between the data domain $X$,
and latent domain $Z$, minimizing the reconstruction objective $d(\delta(\eta(x)), x)$, with a difference measure $d$ (e.g., $L_p$ norm). 
As discussed above, we wish to learn the invariant map $\eta$ ($\varphi_{\text{inv}}$ in the previous paragraph), thus
\begin{property}
\label{prop:eta}
The encoding function $\eta:X\rightarrow Z$ is $G$-invariant, i.e., $\eta(\rho_X(g)x)=\eta(x) \; \forall x \in X, \forall g\in G$.
\end{property}
The decoding function $\delta$ maps the $G$-invariant representation $z\in Z$ back to the data domain $X$. However, as $z$ is $G$-invariant, $\delta$ can at best map $\eta(x)\in Z$ back to an element $\hat{x}\in X$ such that $\hat{x}\in \{\rho_X(g)x|\forall g \in G\}$, i.e., an element in the orbit of $x$ through $G$. 
This is depicted in Figure \ref{fig:main}a.
Thus, the task of the decoding function $\delta:Z\rightarrow X$ is to map encoded elements $z=\eta(x)\in Z$ to an element $\hat{x}\in X$ such that 
$\exists \hat{g}_x\in G$ such that 
\begin{equation}
\delta(\eta(x)) = \hat{x} = \rho_X(\hat{g}_x)x ~.
\end{equation}
We call $\hat{x}$ the \textbf{canonical} element of the decoder $\delta$.
We can rewrite the reconstruction objective with a $G$-invariant encoding function $\eta$ as $d(\rho_X(\hat{g}^{-1})\delta(\eta(x)), x)$. 
One of the main results of this work consists in showing that $\hat{x}$ and $\hat{g}_x$ can be 
\textit{simultaneously} learned by a suitable neural network. That is, we have the following property of our learning scheme:
\begin{property}
\label{prop:definingpsi}
There exists a \textbf{learnable} function $\psi:X\rightarrow G$ such that, given suitable $\eta,\delta$ as described above the relation $ \rho_X(\psi(x))\delta(\eta(x)) = x~, $ holds for all $x \in X$.
\end{property}
We call any function $\psi$ satisfying \eqref{prop:definingpsi} a \textbf{suitable} group function. 
Figure \ref{fig:main}b describes schematically our proposed framework.  
In what follows, we will
first characterize the defining properties of suitable group functions. Subsequently, we will describe our construction, valid for any group $G$.

\subsection{Predicting Group Actions}
In the following we further characterize the properties of $\psi$. We begin by stating two key results, while we refer to the Appendix \ref{app:proofs} for the proofs. 
\begin{proposition}
\label{prop:psiequiv}
Any suitable group function $\psi: X \rightarrow G$ is $G$-equivariant at a point $x\in X$ up the stabilizer $G_x$, i.e., 
$\psi(\rho_X(g)x) \subseteq g\cdot \psi(x) G_x$.
\end{proposition}

\begin{proposition}
\label{prop:surj}
The image of any suitable group function $\psi:X\rightarrow G$ is surjective into $\frac{G}{G_X}$, where $G_X$ is the stabilizers of all the
points of $X$.
\end{proposition}

Let us briefly discuss an example. Suppose $X=\{ x=(x_0,x_1,x_2,x_3) \in \R^{4\times2} | x_i = \rho_{\R^2}(g_{\theta=\pi/2})^i x_0,\ x_0 \in \R^2\}$ and $G=\SO(2)$. $X$ describes all collections of vertices of squares centered at the origin of $\R^2$, and it is easy to check that $g_X = \Z_4$, generated by a $\pi/2$ rotation around the origin. In this case, any such square can be brought to any other square (of the same radius) 
by a rotation of an angle $\theta< \pi/2$, thus $\Imag \psi \supseteq \{ g_\theta\in\SO(2) | 0\leq \theta\leq \pi/2 \} = \SO(2)/\Z_4$. 

Combining the two propositions above we have the following
\begin{lemma}
\label{lem:isom}
Any suitable group function $\psi$ is an isomorphism $O_x \simeq G/G_x$ for any $x\in X$, 
where $O_x\subset X$
is the orbit of $x$ with respect to $G$ in $X$.
\end{lemma}

\subsection{Proposed Construction}
\label{s:proposed_construction}

Next, we turn to our proposed construction of a class of suitable group functions that
satisfy Property \ref{prop:definingpsi} for any data space $X$ and group $G$. 
As we described above, these functions must be learnable.

\begin{property}[\textbf{Proposed construction}]
\label{prop:psi2}
Without loss of generality,
we write our target function $\psi = \xi \circ \mu$, where $\mu: X \rightarrow Y$ is a
\textit{learnable} map between the data space $X$ and the embedding space $Y$, while 
$\xi: Y \rightarrow G$ is a
\textit{deterministic} map. 
Our construction is further determined by the following properties:
\begin{itemize}
    \item We impose $\mu: X \rightarrow Y$ to be $G$-equivariant, that is, $\mu(\rho_X(g)x) = \rho_Y(g)\mu(x)$ for all $x\in X$ and $g\in G$.
    \item We ask that $Y$ is an homogeneous space, that is, given any element $y_0\in Y$, 
    every element $y\in Y$ can be written as $y = \rho_Y(g)y_0$ for some $g\in G$.
    \item The map $\xi: Y \rightarrow G$ is defined as follows: $\xi(y)=g$ such that $y=\rho_Y(g)y_0$ for \textit{any} chosen point $y_0\in Y$.
\end{itemize}
\end{property}
In what follows we will show 
that our construction satisfies the properties of the previous section. For proofs see Appendix.
We begin with the following 
\begin{proposition}
\label{prop:muequiv}
Let $\psi = \xi \circ \mu$ be a suitable group function and let $\mu: X \rightarrow Y$ be $G$-equivariant. Then, $G_x = G_{\mu(x)}$ for all $x\in X$.
\end{proposition}

The result of the above proposition is crucial for our desired decomposition of the 
learned embedding, as it ensures that no information about the group action on $X$ is
lost through the map $\mu$: if a group element acts non-trivially in $X$, it will also act non-trivially in $Y$.

\begin{proposition}
\label{prop:xiequiv}
Given $y, y_0$, the element $g$ such that $y \equiv \rho_Y(g) y_0$ is unique up to the stabilizer $G_{y_0}$.
\end{proposition}

This proposition establishes the equivariant properties of the map $\xi$. Finally, we have
\begin{proposition}
Let $\psi = \xi \circ \mu$ where $\mu$ and $\xi$ are as described above. Then, 
$\psi$ is a suitable group function.
\end{proposition}

\subsection{Intuition Behind the Proposed Framework}

We conclude this rather technical section with a comment on the intuition behind our construction. Assuming for simplicity that the domain set $V$ admits the structure of vector space, $Y$ represents the space spanned by \textbf{all} basis vectors of $V$. The point $y_0$
represent a canonical orientation of such basis, and the element $\xi(y)=g$ is the group element 
corresponding to a basis transformation. As all elements can be expressed in terms of coordinates with respect to a given basis, it is natural to consider a canonical basis for all orbits, justifying the assumption of
homonogeneity of the space $Y$.


Further,
let us assume that the
invariant autoencoder correctly solves its task,
$x \sim \delta(\eta(x))$. 
Now let $\hat{x}\in O_x$ such that $\hat{x}=\delta(\eta(x))$, and by definition, 
$\hat{x}=\rho_V(g)x$ for some $g\in G$. 
Now, the correct orbit element is identified when $\psi(\hat{x})=e$, since 
$\psi(x)=g^{-1}\cdot \psi(\hat{x}) = g^{-1}$ and thus $\rho_X(g^{-1})\delta(\eta(x)) = 
\rho_X(g^{-1}) \hat{x} = x$. 
Hence, during training $\psi$ needs to learn which orbit elements
are decoded as ``canonical'', i.e., without the need of an additional group transformation.
To clarify, here ``canonical'' does not reflect any specific property of the element, but it simply
refers to the orientation learned from the decoder during training. In fact, different decoder
architectures or initializations will lead to different canonical elements.

Finally, note how the different parts of our proposed framework ($\eta$, $\delta$ and $\psi$), as visualized in Figure \ref{fig:main}b, can be jointly trained by minimizing the objective
\begin{equation}
    d(\rho_X(\psi(x))\delta(\eta(x)), x),
\end{equation}
which is \emph{by construction} group invariant, i.e., not susceptible to potential group-related bias in the data (e.g., data that only occurs in certain orientations).

\section{Application to Common Groups}
\label{examples}
In this section we describe how our framework applies to a variety of common groups which we will then implement in our experiments.
As discussed in Section \ref{s:problem} and visualized in Figure \ref{fig:main}b, the main components of our proposed framework are the encoding function $\eta$, the decoding function $\delta$ and the group function $\psi$. As stated in Property \ref{prop:eta}, the only constraint for the encoding function $\eta$ is that it has to be group invariant.
This is in general straightforward to achieve for different groups as we will demonstrate in Section \ref{s:experiments}.
Our proposed framework does not constrain the decoding function $\delta$ other than that it has to map elements from the latent space $Z$ to the data domain $X$. Hence, $\delta$ can be designed independently of the group of interest.
The main challenge is in defining the group function $\psi=\xi\circ\mu$ such that it satisfies Property \ref{prop:definingpsi}. Following Property \ref{prop:psi2} we now turn to describing our
construction of $\xi$, $\mu$ and $Y$ for a variety of common groups.

\paragraph{Orthogonal group $\SO(2)$.}
The Lie group $\SO(2)$ is defined as the set of all rotations about
the origin in $\R^2$.
We take $Y$ to be the circle $S^1\subset \R^2$, that is, the space spanned by unit vectors in $\R^2$. 
Now, $S^1$ is a homogeneous space: any two points $s_0, s_1\in S^1$ are related by a rotation. 
Without loss of generality, we take the reference vector $y_0$ to be the vector $(1,0)\in S^1$. 
Then given a vector $y\in S^1$, we can write
\begin{align}
    y = \begin{pmatrix}
    y_x \\ y_y
    \end{pmatrix}
    =
    \begin{pmatrix}
    y_x & -y_y\\
    y_y & y_x
    \end{pmatrix}
    \begin{pmatrix}
    1 \\ 0
    \end{pmatrix}~.
\end{align}
thus, the function $\xi:S^1\rightarrow \SO(2)$ is determined by $\xi(y) = g_\theta$ such that $\theta = \arccos(y_x) = \arcsin(y_y)$.
\paragraph{Orthogonal group $\SO(3)$.}
We assume that $X$ has no fixed points, as this is usually the case for generic shapes (point clouds) in $\R^3$. 
It would be tempting to take $Y$ to be the sphere $S^2\subset \R^3$, that is, the space spanned by unit vectors in $\R^3$. 
While this space is homogeneous, it does not satisfy the condition that the stabilizers of $G$ are trivial. In fact, given 
any vector $y_1\in S^2$, we have $G_{y_1} = \{ g\in \SO(3) | g \text{ is a rotation about } y_1 \} $~.

In order to construct a space with the desired property, consider a second vector $y_2\in S^2$
orthogonal to $y_1$, $y_2\subset y_1^\perp$. 
Taking $Y$ to be the space spanned by $y_1,y_2\in S^2$, 
it is easy to see that now all the stabilizers are trivial.
Finally, let $y_3 = y_1 \times y_2\in S^2$, then we construct the rotation matrix
\begin{equation*}
R = 
    \begin{pmatrix}
    y_{1,x} & y_{2,x} & y_{3,x}\\
    y_{1,y} & y_{2,y} & y_{3,y}\\
    y_{1,z} & y_{2,z} & y_{3,z}
    \end{pmatrix}~,
\text{which satisfies}~
    \begin{pmatrix}
        y_1 \\ y_2
    \end{pmatrix} = R \begin{pmatrix}
    1 & 0 & 0 \\
    0 & 1 & 0
    \end{pmatrix}^\intercal = R 
    y_0~.
\end{equation*}
    
\paragraph{Symmetric group $S_n$.}

A suitable space $Y$ is the set of ordered collections of unique elements of the set 
$M = \{1,2, \dots, n \}$. For instance, for $n=3$,
we have $Y=\{ (1,2,3), (1,3,2), (2,1,3), (2,3,1), (3,1,2), (3,2,1)\}$. 
It is trivial to see that the action of the permutation group on the set $Y$ is free, that is, all the stabilizers are trivial. 
Explicitely, given any permutation-equivariant vector $w\in \R^n$, we obtain 
an element $y = \text{argsort}(w) \in Y$.
Moreover, it is also obvious that any element in $Y$ can be written as 
$P_{\sigma}(1,2,\dots, n) = (\sigma(1), \sigma(2), \dots, \sigma(n))~,$
that is, a group element acting on the canonical $y_0=(1,2,\dots,n)$. 

\paragraph{Translation group $T_n$.}

Here we take $Y=\R^n$, which is homogeneous with respect to the translation group. In fact, any vector $y\in Y$ can be trivially written as $y=y+ \mathbf{0} = y + y_0$, where $\mathbf{0}$ is the origin of $\R^n$. Our group function takes therefore the form $\xi(y) = y$.

\paragraph{Euclidean group $\SE(n)$.}

A generic transformation of the Euclidean group on a $n$-dimensional representation $v\in V$ is
\begin{align}
    v \mapsto Av + b~, \quad A\in \SO(n)~, b\in T_n~.
\end{align}
Let $\mu = (\mu_1, \mu_2, \dots, \mu_{n+1})$ be a collection of $n+1$ $n$-dimensional $\SE(n)$-equivariant vectors, that is, $\mu_i(\rho_X(g)x)=\rho_Y(g)\mu(x)$, $i=1,\dots,n$.
We construct $\widehat{y}_a = (\mu_a - \mu_{n+1}) / ||\mu_a - \mu_{n+1}||\in S^n$, $a=1,\dots,n$,
where $S^n$ is the unit $n$-dimensional sphere. These 
$n$ ortho-normal vectors are 
\textbf{translation invariant} but rotation equivariant,
and are suitable to construct the rotation matrix
\begin{align}
\label{eq:rotason}
R = 
    \begin{pmatrix}
        \widehat{y}_1 & \widehat{y}_2 & \cdots & \widehat{y}_n
    \end{pmatrix}~,
\end{align}
while the extra vector $\widehat{y}_{n+1} = \mu_{n+1}$ can be used to predict the translation action.
Putting all together, the space $Y$ is described by $n$ vectors 
$y_a = \widehat{y}_a+\widehat{y}_{n+1}$,
and $y_0=I_n$ is the $n\times n$ unit matrix, as
\begin{align}
\label{eq:translation-vector-se(n)}
    (R + \widehat{y}_{n+1})I_n = \begin{pmatrix}
        \widehat{y}_1 & \cdots & \widehat{y}_n
    \end{pmatrix}^\intercal + \widehat{y}_{n+1}I_n =
    \begin{pmatrix}
        y_1 & \cdots & y_n
    \end{pmatrix}^\intercal~.
\end{align}


\section{Related Work}
\paragraph{Group equivariant neural networks.} 

Group equivariant neural networks have shown great success for various groups and data types.
There are two main approaches to implement equivariance in a layer and, hence, in a neural network.
The first, and perhaps the most common, imposes equivariance on the space of functions and features
learned by the network. Thus, the parameters of the model are constrained 
to satisfy equivariance \cite{thomas2018tensor,NEURIPS2019_45d6637b,Weiler20183DSC, Esteves2020SpinWeightedSC}. The disadvantage of this approach consists in the
difficulty of designing suitable architectures for all components of the model, transforming correctly under the group action \cite{xu2021group}.
The second approach to equivariance consists in lifting the map from the 
space of features to the group $G$, and equivariance is defined
on functions on the group itself \cite{Romero2020CoAttentiveEN,Romero2020AttentiveGE,Hoogeboom2018HexaConv}. 
Although this strategy avoids the
architectural constraints, applicability is limited to homogeneous spaces \cite{hutchinson2021lietransformer} and
involves an increased dimensionality of the feature space, due to the lifting to $G$. Equivariance has been explored in a variety of architecture and data structures: Convolutional Neural Networks \cite{cohen2016group, worrall2017harmonic, weiler2018learning,bekkers2018roto, thomas2018tensor, dieleman2016exploiting, kondor2018generalization, cohen2016steerable,cohen2018general, pmlr-v119-finzi20a}, Transformers \cite{vaswani2017attention, fuchs2020se, hutchinson2021lietransformer, romero2020group}, Graph Neural Networks \cite{defferrard2016convolutional, bruna2013spectral, kipf2016semi, gilmer2017neural, satorras2021en} and Normalizing Flows \cite{rezende2015variational, kohler2019equivariant, kohler2020equivariant,boyda2021sampling}. These methods are usually trained in a supervised manner and combined with a symmetric function (e.g. pooling) to extract group-invariant representations.

\paragraph{Group equivariant autoencoders.}
Another line of related work is concerned with group equivariant autoencoders. Such models utilize specific network architectures to encode and decode data in an equivariant way, resulting into equivariant representations only \cite{hinton2011transforming, sabour2017dynamic, kosiorek2019stacked, guo2019affine}.
\citet{feige2019invariantequivariant} use weak supervision in an AE to extract invariant and equivariant representations.
\citet{winter2021permutation} implement a permutation-invariant AE to learn 
graph embeddings, in which the permutation matrix for graph matching is learned during training. 
In that sense, the present work can be seen as a generalization of their approach 
for a generic data type and any group.

\paragraph{Unsupervised invariant representation learning.}
The field of unsupervised invariant representation learning can be roughly divided into two categories. 
The first consists in learning an approximate group action in order to match the input and the reconstructed data. 
For instance, \cite{8579053} propose to encode the input in quotient space, and train the model with a loss that is defined by taking the infimum over the group $G$. While this is feasible for (small) finite groups, for continuous groups they either have to approximately discretize them or perform a separate optimization of the group action at every back propagation step to find the best match. Other work \cite{Shu2018DeformingAU, Koneripalli2020RateInvariantAO} proposes to disentangle the embedding in a shape-like and a deformation-like component. While this is in spirit with our work, their transformations are local (we focus on global transformations) and are approximative, that is, the components are not explicitly invariant and equivariant with respect to the transformation, respectively. 

In the case of 2D/3D data, co-alignment of shapes can be used to match the input and the reconstructed shapes. 
Some approaches are unfeasible \cite{Wang2012ActiveCO} as they are not compatible with a purely unsupervised approach, while other \cite{Averkiou2016AutocorrelationDF, Chaouch2008ANM, Chaouch2009AlignmentO3} leverage symmetry properties of the data and PCA decomposition, exhibiting however limitation regarding scalability. 
For graphs, the problem of graph matching \cite{Bunke2000} has been tackled in several works and with different approaches, for instance algorithmically, e.g., \cite{Ding2020EfficientRG}, or by means of a GNN \cite{Li2019GraphMN}.

On the topic of group theory-based embedding disentanglement, 
some works are based on the definition of \cite{Higgins2018TowardsAD} of a disentangled representations.
We refer to this as ``symmetry-based decomposition'', where the various factors in the disentangled representation 
correspond to the decomposition of symmetry groups acting on the data space. 
In \cite{Pfau2020DisentanglingBS}, the authors show that, with some assumption on the geometry of the
underlying data space,
\begin{wrapfigure}{r}{0.5\textwidth}
     \centering
     \includegraphics[width=0.49 \textwidth]{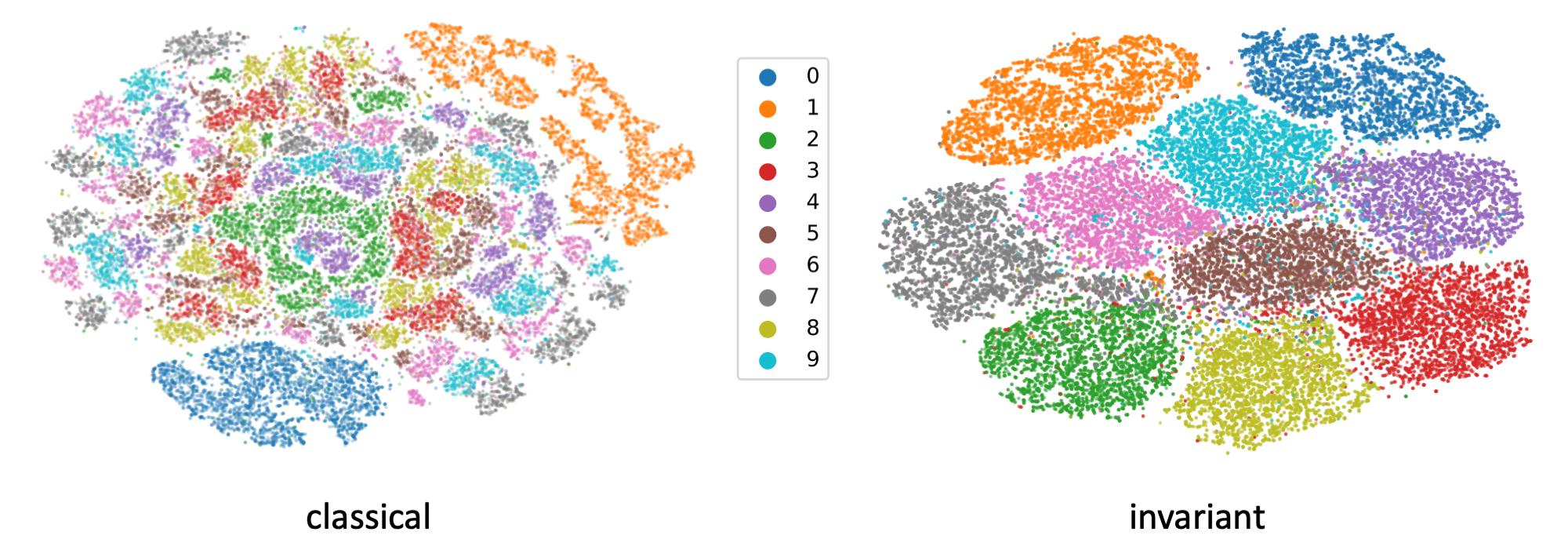}
     \caption{TSNE embedding of the encoded test dataset for a classical and our proposed SO(2) invariant autoencoder.}
     \label{fig:mnist_emb}
     \vspace{-0.15cm}
\end{wrapfigure}
it is possible to learn to factorize a Lie group from the orbits in data space.
The works \cite{Hosoya2019GroupbasedLO, Keurti2022HomomorphismA}, for instance, design unsupervised generative VAEs
approaches for learning representation corresponding to orthogonal symmetry actions on the data space. 
In our work, on the other hand, we learn a decomposition into separate \textit{group representations}. These are 
all representations of the same group, but act differently on different data space (analogously to 
different $\SO(3)$ representations identified by the angular quantum number $l=0,1,2,\dots$).
\section{Experiments}
\label{s:experiments}

In this section we present differnt experiments for the various groups discussed in Section \ref{examples}. \footnote{Source code for the different implementations available at \url{https://github.com/jrwnter/giae}.}
\begin{figure}
     \centering
     \includegraphics[width=0.99\textwidth]{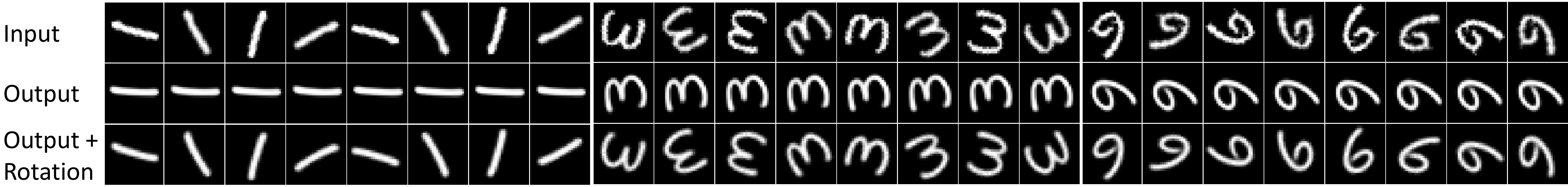}
     \caption{Input and predicted output for rotated versions of three MNIST images. Top row shows the input image successively rotated by $45^{\circ}$. Middle row shows the decoded (canonical) image  and bottom row shows the decoded image after applying the predicted rotation.}
     \label{fig:mnist}
\end{figure}

\subsection{Rotated MNIST}
In the first experiment, we train an SO(2)-invariant autoencoder on the original (non-rotated) MNIST dataset and validate the trained model on the rotated MNIST dataset (ref. \cite{mnistrot}) which consists of randomly rotated versions of the original MNIST dataset. For the functions $\eta$ and $\psi$ we utilize SO(2)-Steerable Convolutional Neural Networks \cite{weiler2019general}. For more details about the network architecture and training, we refer to Appendix \ref{app:mnist}. In Figure \ref{fig:mnist} we show images in different rotations and the respective reconstructed images by the trained model. The model decodes the different rotated versions of the same image (i.e., elements from the same orbit) to the same canonical output orientation (second row in Figure \ref{fig:mnist}). The trained model manages to predict the right rotation matrix (group action) to align the decoded image with the input image, resulting in an overall low reconstruction error.
Note that the model never saw rotated images during training but still manages to encode and reconstruct them due to its inherent equivariant design.
We find that the encoded latent representation is indeed rotation invariant (up to machine precision), but only for rotations of an angle $\theta=\frac{n\cdot\pi}{2},\  n\in \mathbb{N}$.
For all other rotations, we see slight variations in the latent code, which, however, is to be expected due to interpolation artifacts for rotations on a discretized grid. Still, inspecting the 2d-projection of the latent code of our proposed model in Figure \ref{fig:mnist_emb}, we see distinct clusters for each digit class for the different images from the test dataset, independent of the orientation of the digits in the images. In contrast, the latent code of a classical autoencoder exhibits multiple clusters for different orientations of the same digit class. 

\begin{figure}[t!]
     \centering
     \includegraphics[width=0.99\textwidth]{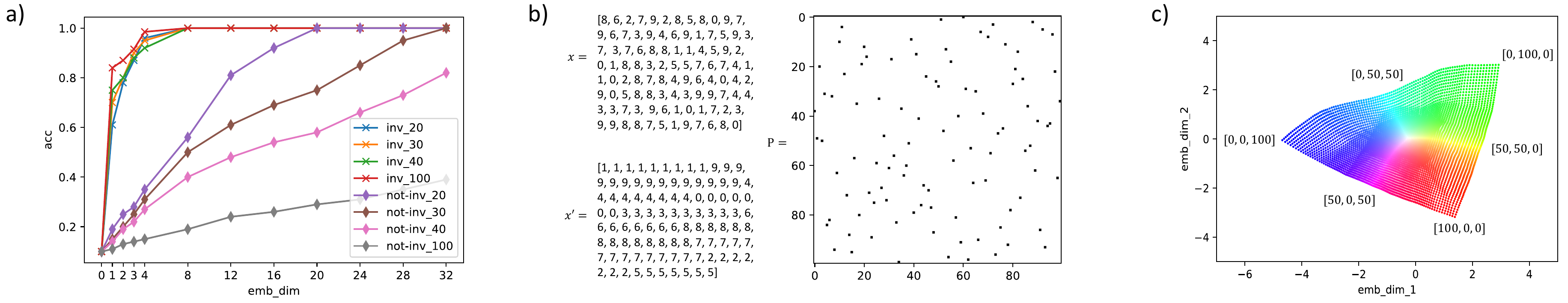}
     \caption{a) Element-wise reconstruction accuracy of our proposed permutation invariant autoencoder (cross) and a classical non-permutation invariant autoencoder (diamond) for different embedding and set sizes. b) Example set $x$ with 100 elements with its canonical reconstruction $\hat{x}$ and the predicted permutation matrix $P$ (resulting into a perfect reconstruction). One can confirm for oneself that, e.g., $x[38] = x'[0]$, matching $P[38, 0]=1$. c) Best viewed in colour. Visualization of the two-dimensional embedding of a permutation-invariant autoencoder for all 5151 sets of 100 elements with 3 different element classes. Each point represents one set, colours represent set compositions (proportion of each element class, independent of the order).}
     \label{fig:digits}
     \vspace{-0.15cm}
\end{figure}

\subsection{Set of Digits}
Next, we train a permutation-invariant autoencoder on sets of digits. A set with $N$ digits is represented by concatenating one-hot vectors of each digit in a $N\times D$-dimensional matrix, where we take $D=10$. Notice that this matrix-representation of a set is \emph{not} permutation invariant. We randomly sampled 1.000.000 different sets for training and 100.000 for the final evaluation with $N=20, 30, 40, 100$, respectively, removing all permutation equivariant sets (i.e., there are no two sets that are the same up to a permutation). For comparison, we additionally trained a classical non-permutation-invariant autoencoder with the same number of parameters and layers as our permutation-invariant version. For more details on the network architecture and training we refer to Appendix \ref{app:digits}. Here, we demonstrate how the separation of the permutation-invariant information of the set (i.e., the composition of the set) from the (irrelevant) order-information results in a significant reduction of the space needed to encode the set. In Figure \ref{fig:digits}a, we plot the element-wise reconstruction accuracy of different sized sets for both models for varying embedding (bottleneck) sizes. As the classical autoencoder has to store both the composition of digits in the set (i.e., number of elements for each of the 10 digits classes) as well as their order in the permutation-dependent matrix representation, the reconstruction accuracy drops for increasing size of the set $N$ for a fixed embedding size. For the same reason, perfect reconstruction accuracy is only achieved if the embedding dimension is at least as large as the number of digits in the set. On the contrary, our proposed permutation invariant autoencoder achieves perfect reconstruction accuracy with a significant lower embedding size. Crucially, as no order information has to be stored in the embedding, this embedding size for perfect reconstruction accuracy also stays the same for increasing size $N$ of the set. In Figure \ref{fig:digits}b we show one example for a set $x$ with $N=100$ digits, with the predicted canonical orbit element $\hat{x}$ and the predicted permutation matrix. As perhaps expected, the canonical element clusters together digits with same value, while not using the commonly used order of Arabic numerals. This learned order (here [1,9,4,0,3,6,8,7,2,5]) stays fixed for the trained network for different inputs but changes upon re-initialization of the network.


In Figure \ref{fig:digits}c we show the two-dimensional embedding of a permutation invariant autoencoder trained on set of $N=100$ elements chosen from $D=3$ different classes (e.g. digits 0,1,2). As the sets only consists of 3 different elements (but in different compositions and order) we can visualize the $\binom{D + N -1 }{N} = \binom{102}{100} = 5151$ elements in the two-dimensional embedding and colour them according to their composition. As our proposed auteoncoder only needs to store the information about the set composition and not the order, the embedding is perfectly structured with respect to the composition as can be seen by the colour gradients in the visualization of the embedding.

\subsection{Point Cloud}
\begin{figure}
     \centering
     \includegraphics[width=0.99\textwidth]{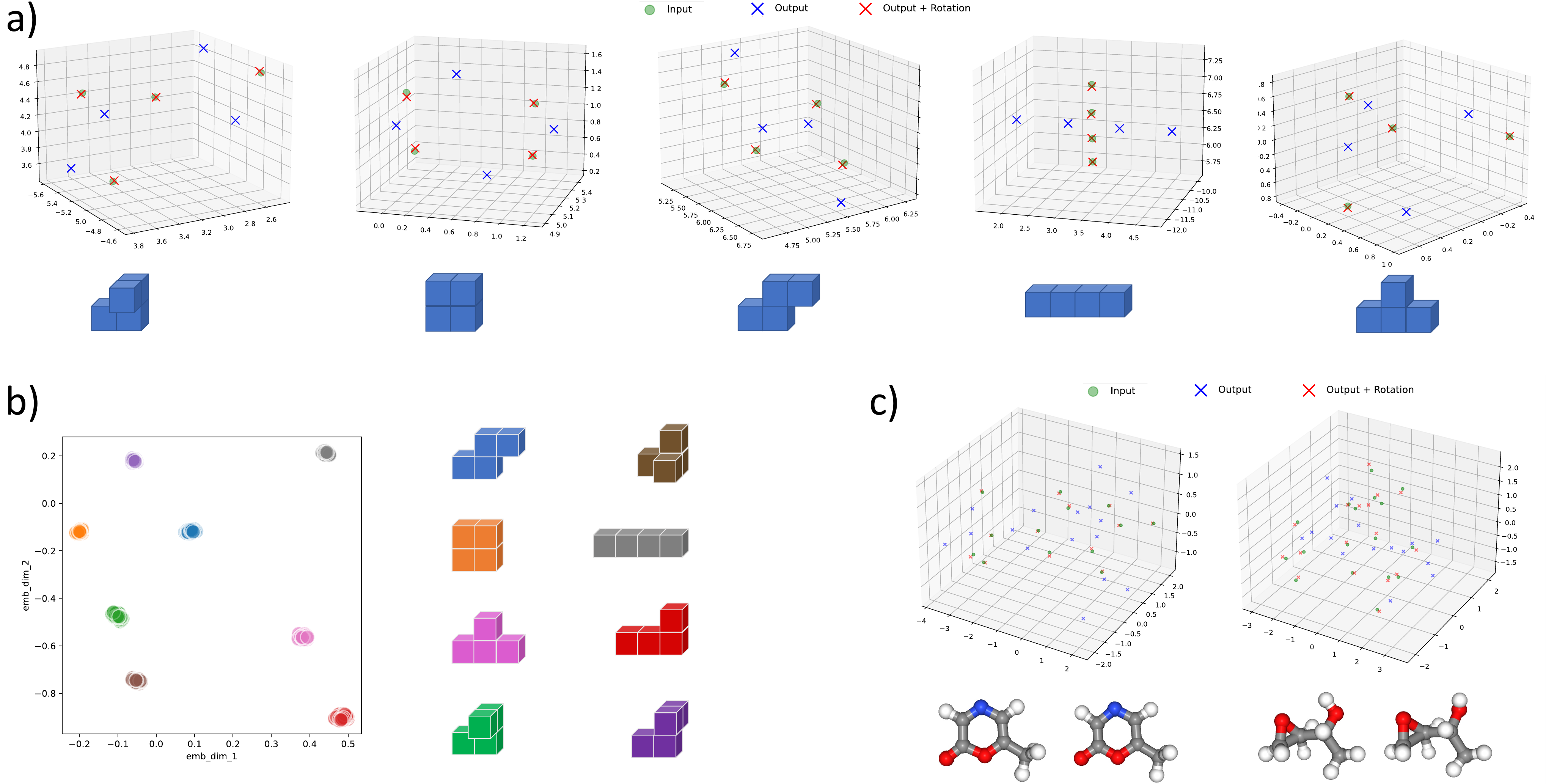}
     \caption{a) Five different Tetris shapes represented by points at the center of the four blocks respectively. Input points, output points and rotated (predicted group action) output points as reconstructed by our proposed SE($3$)- and S($N$)-invariant autoencoder are visualized. b) Two-dimensional latent space for all Tetris shapes augmented with Gaussian noise ($\sigma=0.01$). Colors of points match colors of shapes on the right. c) Two molecular conformations and their reconstructions represented as point cloud and ball-and-stick model (left true, right predicted).\\
     }
     \label{fig:tetris}
     \vspace{-0.15cm}
\end{figure}

Point clouds are a common way to describe objects in 3D space, such as the atom positions of a molecule or the surface of an object. As such, they usually adhere to 3D translation and rotation symmetries and are unordered, i.e., permutation invariant. Hence, we investigate in the next experiment a combined SE($3$)- and $S_N$-invariant autoencoder for point cloud data. We use the Tetris Shape toy dataset \cite{thomas2018tensor} which consists of 8 shapes, where each shape includes $N=4$ points in $3$D space, representing the center of each Tetris block. To generate various shapes, we augment the 8 shapes by adding Gaussian noise with $\sigma=0.01$ standard deviation on each node's position. Different orientations are obtained by rotating the point cloud with a random rotation matrix $R\in \textsc{SO}(3)$ and further translating all node positions with the same random translation vector $t\in \R^{3} \simeq T_3$.
For additional details on the network architecture and training 
we refer to Appendix \ref{app:pointcloud}.
In Figure \ref{fig:tetris} we visualize the input points and output points before and after applying the predicted rotation. The model successfully reconstructs the input points with high fidelity (mean squared error of $\sim 4\times10^{-5}$) for all shapes and arbitrary translations and rotations.
Figure \ref{fig:tetris}b shows the two-dimensional embedding of the trained SE($3$)- and $S_N$-invariant autoencoder. Augmenting the points with random noise results into slight variations in the embedding, while samples of the same Tetris shape class still cluster together. The embedding is invariant with respect to rotations, translation and permutations of the points. Notably, the SE(3)-invariant representations can distinguish the two chiral shapes (compare green and violet coloured shapes in the bottom right of Figure \ref{fig:tetris}b). These two shapes are mirrored versions of themselves and should be distinguished in an SE($3$) equivariant model. Models that achieve SE($3$) invariant representations by restricting themselves to composition of symmetric functions only, such as working solely on distances (e.g. SchNet \cite{schutt2018schnet}) or angles (e.g. ANI-1 \cite{smith2017ani}) between points fail to distinguish these two shapes \cite{thomas2018tensor}.

\paragraph{Molecular Conformations.}
We showcase our learning framework on real-world data by autoencoding the atom types and geometries of small molecules from the \texttt{QM9} database \cite{ramakrishnan2014quantum}.
We achieved a reconstruction RMSE of $0.15 \pm 0.07~\angstrom$ for atom coordinates and perfect atom type accuracy on 5000 unseen test conformations (see Figure \ref{fig:tetris}c for two examples and Appendix \ref{app:qm9examples} for more reconstruction predictions). Given a point cloud of $N$ nodes, the $G=SE(3)\times S_N$-invariant embedding $z$ has to store information about the Cartesian coordinates $P \in \mathbb{R}^{3N}$ as well as the 5 distinct atom types $A \in \{0,1\}^{5N}$ represented as one-hot encodings. The largest molecule in the QM9 database has $N_\text{max} = 29$ atoms, thus the degrees of freedom of the data space\footnote{Notice that the data space $X$ can be described as the product space between $\mathbb{R}^{3N}$ and $\mathbb{N}^{5N}$.} $X$ are $3\cdot29 \cdot 5 \cdot 29 = 12615$.
Our embeddings compress this high-dimensional space of molecular conformations into $z\in Z \subset \mathbb{R}^{256}$ dimensions.
\begin{wrapfigure}{r}{0.5\textwidth}
    \centering
    \includegraphics[width=0.53 \textwidth]{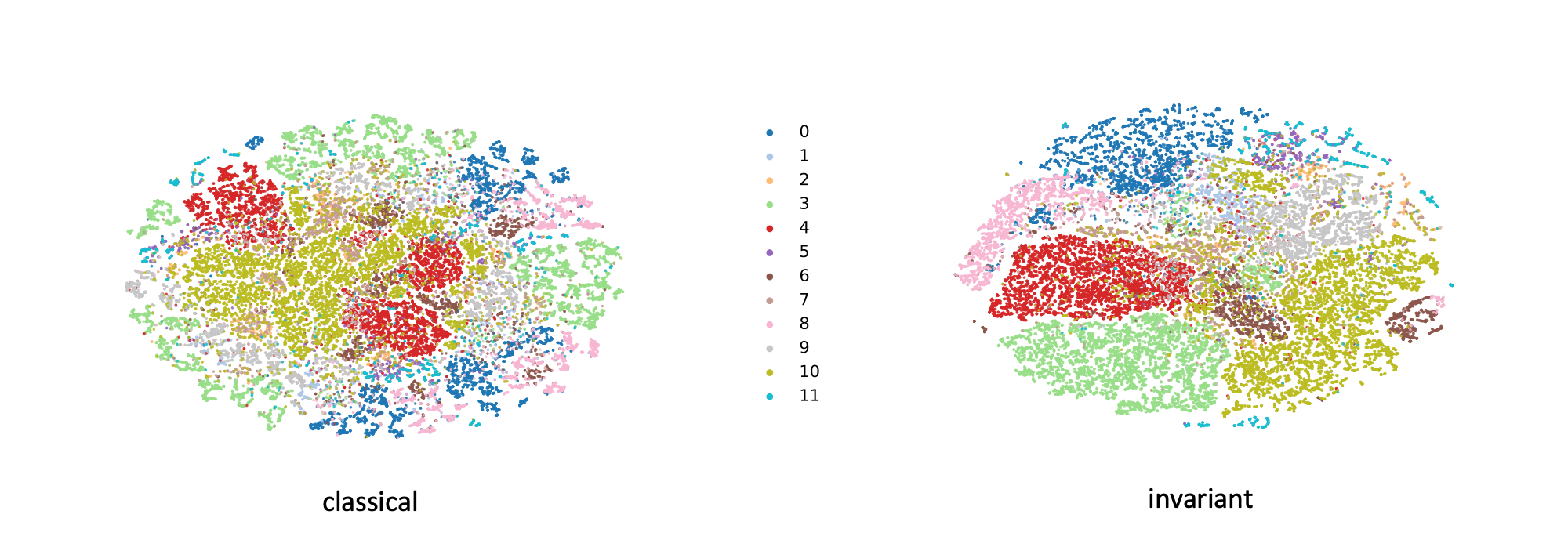}
    \caption{TSNE embedding of the encoded dataset for a classical autoencoder and our proposed SO(3) invariant autoencoder.}
    \label{fig:tsne_shapenet}
    \vspace{-1cm}
\end{wrapfigure} 
\subsection{ShapeNet}
We also run experiments on the ShapeNet dataset \cite{chang2015shapenet}. We utilized 3D Steerable CNNs proposed by \cite{weiler20183d} as equivariant encoder for the 3d voxel input space. We utilized the scalar outputs as rotation-invariant embedding ($z$) and predict (analogously to our experiments on 3d point clouds) 2 rotation-equivariant vectors to construct a rotation matrix $\rho(g)$. In Figure \ref{fig:shapenet_rec} in the Appendix we show example reconstructions of shapes from the SE$(3)$ invariant representations. Similar to our MNIST experiment, we compared the resulting embedding space to the embeddings produced by a non-invariant autoencoder model. 
As the dataset comes in an aligned form (e.g., cars are always aligned in the same orientation), we additionally applied random 90 degree rotations to remove this bias (while avoiding interpolation artifacts) when training the non-invariant model. Random rotations are also applied to the common test set. In Figure \ref{fig:tsne_shapenet} we visualize a TSNE projection of the embeddings of both models. We can see a well structured embedding space for our model with distinct clusters for the different shape classes. On the other hand, the embeddings produced by the non-invariant autoencoder is less structured and one can make out different clusters for the same shape label but in different orientations. Moreover, we compared the downstream performance and generalizability of a KNN classifier on shape classification, trained on 1000 embeddings and tested on the rest. The classifier based on our rotation-invariant embeddings achieved an accuracy of 0.81 while the classifier based on the non-invariant embeddings achieved an accuracy of only 0.63.

\section{Conclusion and Future Work}
In this work we proposed a novel unsupervised learning strategy to extract representations from data that are separated in a group invariant and equivariant part for any group $G$. We defined the sufficient conditions for the different parts of our proposed framework, namely the encoder, decoder and group function without further constraining the choice of a ($G$-) specific network architecture. In fact, we demonstrate the validity and flexibility of our proposed framework for diverse data types, groups and network architectures. 

To the best of our knowledge, we propose the first general framework for unsupervised learning of separated invariant-equivariant representations valid for any group. Our learning strategy can be applied to any AE framework,
including variational AEs. It would be compelling to extend our approach to a fully probabilistic approach, where the group action function samples from a probability distribution. Such formalism would be relevant in scenarios where some elements of a group orbit occur with different frequencies, enabling this to be reflected in the generation process. For instance, 
predicting protein-ligand binding sites depends on the molecule's orientation with 
respect to the protein pocket or cavity. Thus, in a generative approach, it would be highly compelling to generate a group action
reflecting a candidate molecule's orientation in addition to a candidate ligand. We plan to return to these generalization and apply our learning strategy to non-trivial real-world applications in future work.

\clearpage
\bibliography{main2.bib}

\begin{thebibliography}{71}
\providecommand{\natexlab}[1]{#1}
\providecommand{\url}[1]{\texttt{#1}}
\expandafter\ifx\csname urlstyle\endcsname\relax
  \providecommand{\doi}[1]{doi: #1}\else
  \providecommand{\doi}{doi: \begingroup \urlstyle{rm}\Url}\fi

\bibitem[mni()]{mnistrot}
{Rotated MNIST}.
\newblock
  \url{https://sites.google.com/a/lisa.iro.umontreal.ca/public_static_twiki/variations-on-the-mnist-digits}.
\newblock [Online; accessed 05-January-2021].

\bibitem[Anderson et~al.(2019)Anderson, Hy, and Kondor]{anderson2019cormorant}
Anderson, B., Hy, T.-S., and Kondor, R.
\newblock Cormorant: Covariant molecular neural networks.
\newblock \emph{arXiv preprint arXiv:1906.04015}, 2019.

\bibitem[Averkiou et~al.(2016)Averkiou, Kim, and
  Mitra]{Averkiou2016AutocorrelationDF}
Averkiou, M., Kim, V.~G., and Mitra, N.~J.
\newblock Autocorrelation descriptor for efficient co‐alignment of 3d shape
  collections.
\newblock \emph{Computer Graphics Forum}, 35, 2016.

\bibitem[Axelrod \& Gomez-Bombarelli(2021)Axelrod and
  Gomez-Bombarelli]{geom-axelrod}
Axelrod, S. and Gomez-Bombarelli, R.
\newblock {GEOM}, 2021.
\newblock URL \url{https://doi.org/10.7910/DVN/JNGTDF}.

\bibitem[Bekkers et~al.(2018)Bekkers, Lafarge, Veta, Eppenhof, Pluim, and
  Duits]{bekkers2018roto}
Bekkers, E.~J., Lafarge, M.~W., Veta, M., Eppenhof, K.~A., Pluim, J.~P., and
  Duits, R.
\newblock Roto-translation covariant convolutional networks for medical image
  analysis.
\newblock In \emph{International conference on medical image computing and
  computer-assisted intervention}, pp.\  440--448. Springer, 2018.

\bibitem[Boyda et~al.(2021)Boyda, Kanwar, Racani{\`e}re, Rezende, Albergo,
  Cranmer, Hackett, and Shanahan]{boyda2021sampling}
Boyda, D., Kanwar, G., Racani{\`e}re, S., Rezende, D.~J., Albergo, M.~S.,
  Cranmer, K., Hackett, D.~C., and Shanahan, P.~E.
\newblock Sampling using su (n) gauge equivariant flows.
\newblock \emph{Physical Review D}, 103\penalty0 (7):\penalty0 074504, 2021.

\bibitem[Bronstein et~al.(2021)Bronstein, Bruna, Cohen, and
  Veli{\v{c}}kovi{\'c}]{bronstein2021geometric}
Bronstein, M.~M., Bruna, J., Cohen, T., and Veli{\v{c}}kovi{\'c}, P.
\newblock Geometric deep learning: Grids, groups, graphs, geodesics, and
  gauges.
\newblock \emph{arXiv preprint arXiv:2104.13478}, 2021.

\bibitem[Bruna et~al.(2013)Bruna, Zaremba, Szlam, and LeCun]{bruna2013spectral}
Bruna, J., Zaremba, W., Szlam, A., and LeCun, Y.
\newblock Spectral networks and locally connected networks on graphs.
\newblock \emph{arXiv preprint arXiv:1312.6203}, 2013.

\bibitem[Bunke \& Jiang(2000)Bunke and Jiang]{Bunke2000}
Bunke, H. and Jiang, X.
\newblock \emph{Graph Matching and Similarity}, pp.\  281--304.
\newblock Springer US, Boston, MA, 2000.
\newblock ISBN 978-1-4615-4401-2.
\newblock \doi{10.1007/978-1-4615-4401-2_10}.
\newblock URL \url{https://doi.org/10.1007/978-1-4615-4401-2_10}.

\bibitem[Chang et~al.(2015)Chang, Funkhouser, Guibas, Hanrahan, Huang, Li,
  Savarese, Savva, Song, Su, et~al.]{chang2015shapenet}
Chang, A.~X., Funkhouser, T., Guibas, L., Hanrahan, P., Huang, Q., Li, Z.,
  Savarese, S., Savva, M., Song, S., Su, H., et~al.
\newblock Shapenet: An information-rich 3d model repository.
\newblock \emph{arXiv preprint arXiv:1512.03012}, 2015.

\bibitem[Chaouch \& Verroust-Blondet(2008)Chaouch and
  Verroust-Blondet]{Chaouch2008ANM}
Chaouch, M. and Verroust-Blondet, A.
\newblock A novel method for alignment of 3d models.
\newblock \emph{2008 IEEE International Conference on Shape Modeling and
  Applications}, pp.\  187--195, 2008.

\bibitem[Chaouch \& Verroust-Blondet(2009)Chaouch and
  Verroust-Blondet]{Chaouch2009AlignmentO3}
Chaouch, M. and Verroust-Blondet, A.
\newblock Alignment of 3d models.
\newblock \emph{Graph. Model.}, 71:\penalty0 63--76, 2009.

\bibitem[Cohen \& Welling(2016{\natexlab{a}})Cohen and Welling]{cohen2016group}
Cohen, T. and Welling, M.
\newblock Group equivariant convolutional networks.
\newblock In \emph{International conference on machine learning}, pp.\
  2990--2999. PMLR, 2016{\natexlab{a}}.

\bibitem[Cohen et~al.(2018)Cohen, Geiger, and Weiler]{cohen2018general}
Cohen, T., Geiger, M., and Weiler, M.
\newblock A general theory of equivariant cnns on homogeneous spaces.
\newblock \emph{arXiv preprint arXiv:1811.02017}, 2018.

\bibitem[Cohen \& Welling(2016{\natexlab{b}})Cohen and
  Welling]{cohen2016steerable}
Cohen, T.~S. and Welling, M.
\newblock Steerable cnns.
\newblock \emph{arXiv preprint arXiv:1612.08498}, 2016{\natexlab{b}}.

\bibitem[Defferrard et~al.(2016)Defferrard, Bresson, and
  Vandergheynst]{defferrard2016convolutional}
Defferrard, M., Bresson, X., and Vandergheynst, P.
\newblock Convolutional neural networks on graphs with fast localized spectral
  filtering.
\newblock \emph{Advances in neural information processing systems},
  29:\penalty0 3844--3852, 2016.

\bibitem[Dieleman et~al.(2016)Dieleman, De~Fauw, and
  Kavukcuoglu]{dieleman2016exploiting}
Dieleman, S., De~Fauw, J., and Kavukcuoglu, K.
\newblock Exploiting cyclic symmetry in convolutional neural networks.
\newblock In \emph{International conference on machine learning}, pp.\
  1889--1898. PMLR, 2016.

\bibitem[Ding et~al.(2020)Ding, Ma, Wu, and Xu]{Ding2020EfficientRG}
Ding, J., Ma, Z., Wu, Y., and Xu, J.
\newblock Efficient random graph matching via degree profiles.
\newblock \emph{Probability Theory and Related Fields}, 179:\penalty0 29--115,
  2020.

\bibitem[Esteves et~al.(2020)Esteves, Makadia, and
  Daniilidis]{Esteves2020SpinWeightedSC}
Esteves, C., Makadia, A., and Daniilidis, K.
\newblock Spin-weighted spherical cnns.
\newblock \emph{ArXiv}, abs/2006.10731, 2020.

\bibitem[Feige(2019)]{feige2019invariantequivariant}
Feige, I.
\newblock Invariant-equivariant representation learning for multi-class data,
  2019.

\bibitem[Finzi et~al.(2020)Finzi, Stanton, Izmailov, and
  Wilson]{pmlr-v119-finzi20a}
Finzi, M., Stanton, S., Izmailov, P., and Wilson, A.~G.
\newblock Generalizing convolutional neural networks for equivariance to lie
  groups on arbitrary continuous data.
\newblock In III, H.~D. and Singh, A. (eds.), \emph{Proceedings of the 37th
  International Conference on Machine Learning}, volume 119 of
  \emph{Proceedings of Machine Learning Research}, pp.\  3165--3176. PMLR,
  13--18 Jul 2020.
\newblock URL \url{https://proceedings.mlr.press/v119/finzi20a.html}.

\bibitem[Fuchs et~al.(2020)Fuchs, Worrall, Fischer, and Welling]{fuchs2020se}
Fuchs, F.~B., Worrall, D.~E., Fischer, V., and Welling, M.
\newblock Se (3)-transformers: 3d roto-translation equivariant attention
  networks.
\newblock \emph{arXiv preprint arXiv:2006.10503}, 2020.

\bibitem[Gilmer et~al.(2017)Gilmer, Schoenholz, Riley, Vinyals, and
  Dahl]{gilmer2017neural}
Gilmer, J., Schoenholz, S.~S., Riley, P.~F., Vinyals, O., and Dahl, G.~E.
\newblock Neural message passing for quantum chemistry.
\newblock In \emph{International conference on machine learning}, pp.\
  1263--1272. PMLR, 2017.

\bibitem[Grover et~al.(2019)Grover, Wang, Zweig, and
  Ermon]{grover2019stochastic}
Grover, A., Wang, E., Zweig, A., and Ermon, S.
\newblock Stochastic optimization of sorting networks via continuous
  relaxations.
\newblock \emph{arXiv preprint arXiv:1903.08850}, 2019.

\bibitem[Guo et~al.(2019)Guo, Zhu, Liu, and Yin]{guo2019affine}
Guo, X., Zhu, E., Liu, X., and Yin, J.
\newblock Affine equivariant autoencoder.
\newblock In \emph{IJCAI}, pp.\  2413--2419, 2019.

\bibitem[Higgins et~al.(2018)Higgins, Amos, Pfau, Racani{\`e}re, Matthey,
  Rezende, and Lerchner]{Higgins2018TowardsAD}
Higgins, I., Amos, D., Pfau, D., Racani{\`e}re, S., Matthey, L., Rezende,
  D.~J., and Lerchner, A.
\newblock Towards a definition of disentangled representations.
\newblock \emph{ArXiv}, abs/1812.02230, 2018.

\bibitem[Hinton et~al.(2011)Hinton, Krizhevsky, and
  Wang]{hinton2011transforming}
Hinton, G.~E., Krizhevsky, A., and Wang, S.~D.
\newblock Transforming auto-encoders.
\newblock In \emph{International conference on artificial neural networks},
  pp.\  44--51. Springer, 2011.

\bibitem[Hoogeboom et~al.(2018)Hoogeboom, Peters, Cohen, and
  Welling]{Hoogeboom2018HexaConv}
Hoogeboom, E., Peters, J. W.~T., Cohen, T., and Welling, M.
\newblock Hexaconv.
\newblock \emph{ArXiv}, abs/1803.02108, 2018.

\bibitem[Hosoya(2019)]{Hosoya2019GroupbasedLO}
Hosoya, H.
\newblock Group-based learning of disentangled representations with
  generalizability for novel contents.
\newblock In \emph{IJCAI}, 2019.

\bibitem[Hutchinson et~al.(2021)Hutchinson, Le~Lan, Zaidi, Dupont, Teh, and
  Kim]{hutchinson2021lietransformer}
Hutchinson, M.~J., Le~Lan, C., Zaidi, S., Dupont, E., Teh, Y.~W., and Kim, H.
\newblock Lietransformer: Equivariant self-attention for lie groups.
\newblock In \emph{International Conference on Machine Learning}, pp.\
  4533--4543. PMLR, 2021.

\bibitem[Keurti et~al.(2022)Keurti, Pan, Besserve, Grewe, and
  Scholkopf]{Keurti2022HomomorphismA}
Keurti, H., Pan, H.-R., Besserve, M., Grewe, B.~F., and Scholkopf, B.
\newblock Homomorphism autoencoder - learning group structured representations
  from observed transitions.
\newblock \emph{ArXiv}, abs/2207.12067, 2022.

\bibitem[Kipf \& Welling(2016)Kipf and Welling]{kipf2016semi}
Kipf, T.~N. and Welling, M.
\newblock Semi-supervised classification with graph convolutional networks.
\newblock \emph{arXiv preprint arXiv:1609.02907}, 2016.

\bibitem[Klicpera et~al.(2020)Klicpera, Gro{\ss}, and
  G{\"u}nnemann]{klicpera2020directional}
Klicpera, J., Gro{\ss}, J., and G{\"u}nnemann, S.
\newblock Directional message passing for molecular graphs.
\newblock \emph{arXiv preprint arXiv:2003.03123}, 2020.

\bibitem[K{\"o}hler et~al.(2019)K{\"o}hler, Klein, and
  No{\'e}]{kohler2019equivariant}
K{\"o}hler, J., Klein, L., and No{\'e}, F.
\newblock Equivariant flows: sampling configurations for multi-body systems
  with symmetric energies.
\newblock \emph{arXiv preprint arXiv:1910.00753}, 2019.

\bibitem[K{\"o}hler et~al.(2020)K{\"o}hler, Klein, and
  No{\'e}]{kohler2020equivariant}
K{\"o}hler, J., Klein, L., and No{\'e}, F.
\newblock Equivariant flows: exact likelihood generative learning for symmetric
  densities.
\newblock In \emph{International Conference on Machine Learning}, pp.\
  5361--5370. PMLR, 2020.

\bibitem[Kondor \& Trivedi(2018)Kondor and Trivedi]{kondor2018generalization}
Kondor, R. and Trivedi, S.
\newblock On the generalization of equivariance and convolution in neural
  networks to the action of compact groups.
\newblock In \emph{International Conference on Machine Learning}, pp.\
  2747--2755. PMLR, 2018.

\bibitem[Koneripalli et~al.(2020)Koneripalli, Lohit, Anirudh, and
  Turaga]{Koneripalli2020RateInvariantAO}
Koneripalli, K., Lohit, S., Anirudh, R., and Turaga, P.~K.
\newblock Rate-invariant autoencoding of time-series.
\newblock \emph{ICASSP 2020 - 2020 IEEE International Conference on Acoustics,
  Speech and Signal Processing (ICASSP)}, pp.\  3732--3736, 2020.

\bibitem[Kosiorek et~al.(2019)Kosiorek, Sabour, Teh, and
  Hinton]{kosiorek2019stacked}
Kosiorek, A.~R., Sabour, S., Teh, Y.~W., and Hinton, G.~E.
\newblock Stacked capsule autoencoders.
\newblock \emph{arXiv preprint arXiv:1906.06818}, 2019.

\bibitem[LeCun et~al.(1995)LeCun, Bengio, et~al.]{lecun1995convolutional}
LeCun, Y., Bengio, Y., et~al.
\newblock Convolutional networks for images, speech, and time series.
\newblock \emph{The handbook of brain theory and neural networks}, 1995.

\bibitem[Li et~al.(2019)Li, Gu, Dullien, Vinyals, and Kohli]{Li2019GraphMN}
Li, Y., Gu, C., Dullien, T., Vinyals, O., and Kohli, P.
\newblock Graph matching networks for learning the similarity of graph
  structured objects.
\newblock \emph{ArXiv}, abs/1904.12787, 2019.

\bibitem[Lyle et~al.(2020)Lyle, van~der Wilk, Kwiatkowska, Gal, and
  Bloem-Reddy]{Lyle2020OnTB}
Lyle, C., van~der Wilk, M., Kwiatkowska, M.~Z., Gal, Y., and Bloem-Reddy, B.
\newblock On the benefits of invariance in neural networks.
\newblock \emph{ArXiv}, abs/2005.00178, 2020.

\bibitem[Mehr et~al.(2018{\natexlab{a}})Mehr, Lieutier, Bermudez, Guitteny,
  Thome, and Cord]{mehr2018manifold}
Mehr, E., Lieutier, A., Bermudez, F.~S., Guitteny, V., Thome, N., and Cord, M.
\newblock Manifold learning in quotient spaces.
\newblock In \emph{Proceedings of the IEEE Conference on Computer Vision and
  Pattern Recognition}, pp.\  9165--9174, 2018{\natexlab{a}}.

\bibitem[Mehr et~al.(2018{\natexlab{b}})Mehr, Lieutier, Bermudez, Guitteny,
  Thome, and Cord]{8579053}
Mehr, Ã., Lieutier, A., Bermudez, F.~S., Guitteny, V., Thome, N., and Cord, M.
\newblock Manifold learning in quotient spaces.
\newblock In \emph{2018 IEEE/CVF Conference on Computer Vision and Pattern
  Recognition}, pp.\  9165--9174, 2018{\natexlab{b}}.
\newblock \doi{10.1109/CVPR.2018.00955}.

\bibitem[Miller et~al.(2020)Miller, Geiger, Smidt, and
  No{\'e}]{miller2020relevance}
Miller, B.~K., Geiger, M., Smidt, T.~E., and No{\'e}, F.
\newblock Relevance of rotationally equivariant convolutions for predicting
  molecular properties.
\newblock \emph{arXiv preprint arXiv:2008.08461}, 2020.

\bibitem[Pfau et~al.(2020)Pfau, Higgins, Botev, and
  Racani{\`e}re]{Pfau2020DisentanglingBS}
Pfau, D., Higgins, I., Botev, A., and Racani{\`e}re, S.
\newblock Disentangling by subspace diffusion.
\newblock \emph{ArXiv}, abs/2006.12982, 2020.

\bibitem[Prillo \& Eisenschlos(2020)Prillo and Eisenschlos]{prillo2020softsort}
Prillo, S. and Eisenschlos, J.
\newblock Softsort: A continuous relaxation for the argsort operator.
\newblock In \emph{International Conference on Machine Learning}, pp.\
  7793--7802. PMLR, 2020.

\bibitem[Puny et~al.(2021)Puny, Atzmon, Ben-Hamu, Misra, Grover, Smith, and
  Lipman]{Omriframeaveraging}
Puny, O., Atzmon, M., Ben-Hamu, H., Misra, I., Grover, A., Smith, E.~J., and
  Lipman, Y.
\newblock Frame averaging for invariant and equivariant network design, 2021.
\newblock URL \url{https://arxiv.org/abs/2110.03336}.

\bibitem[Ramakrishnan et~al.(2014)Ramakrishnan, Dral, Rupp, and von
  Lilienfeld]{ramakrishnan2014quantum}
Ramakrishnan, R., Dral, P.~O., Rupp, M., and von Lilienfeld, O.~A.
\newblock Quantum chemistry structures and properties of 134 kilo molecules.
\newblock \emph{Scientific Data}, 1, 2014.

\bibitem[Rezende \& Mohamed(2015)Rezende and Mohamed]{rezende2015variational}
Rezende, D. and Mohamed, S.
\newblock Variational inference with normalizing flows.
\newblock In \emph{International conference on machine learning}, pp.\
  1530--1538. PMLR, 2015.

\bibitem[Romero \& Cordonnier(2020)Romero and Cordonnier]{romero2020group}
Romero, D.~W. and Cordonnier, J.-B.
\newblock Group equivariant stand-alone self-attention for vision.
\newblock \emph{arXiv preprint arXiv:2010.00977}, 2020.

\bibitem[Romero \& Hoogendoorn(2020)Romero and
  Hoogendoorn]{Romero2020CoAttentiveEN}
Romero, D.~W. and Hoogendoorn, M.
\newblock Co-attentive equivariant neural networks: Focusing equivariance on
  transformations co-occurring in data.
\newblock \emph{ArXiv}, abs/1911.07849, 2020.

\bibitem[Romero et~al.(2020)Romero, Bekkers, Tomczak, and
  Hoogendoorn]{Romero2020AttentiveGE}
Romero, D.~W., Bekkers, E.~J., Tomczak, J.~M., and Hoogendoorn, M.
\newblock Attentive group equivariant convolutional networks.
\newblock \emph{ArXiv}, abs/2002.03830, 2020.

\bibitem[Sabour et~al.(2017)Sabour, Frosst, and Hinton]{sabour2017dynamic}
Sabour, S., Frosst, N., and Hinton, G.~E.
\newblock Dynamic routing between capsules.
\newblock \emph{arXiv preprint arXiv:1710.09829}, 2017.

\bibitem[Satorras et~al.(2021)Satorras, Hoogeboom, and Welling]{satorras2021en}
Satorras, V.~G., Hoogeboom, E., and Welling, M.
\newblock E(n) equivariant graph neural networks, 2021.

\bibitem[Sch{\"u}tt et~al.(2018)Sch{\"u}tt, Sauceda, Kindermans, Tkatchenko,
  and M{\"u}ller]{schutt2018schnet}
Sch{\"u}tt, K.~T., Sauceda, H.~E., Kindermans, P.-J., Tkatchenko, A., and
  M{\"u}ller, K.-R.
\newblock Schnet--a deep learning architecture for molecules and materials.
\newblock \emph{The Journal of Chemical Physics}, 148\penalty0 (24):\penalty0
  241722, 2018.

\bibitem[Sch\"{u}tt et~al.(2021)Sch\"{u}tt, Unke, and
  Gastegger]{schuett2021equivariant}
Sch\"{u}tt, K.~T., Unke, O.~T., and Gastegger, M.
\newblock Equivariant message passing for the prediction of tensorial
  properties and molecular spectra, 2021.

\bibitem[Shu et~al.(2018)Shu, Sahasrabudhe, G{\"u}ler, Samaras, Paragios, and
  Kokkinos]{Shu2018DeformingAU}
Shu, Z., Sahasrabudhe, M., G{\"u}ler, R.~A., Samaras, D., Paragios, N., and
  Kokkinos, I.
\newblock Deforming autoencoders: Unsupervised disentangling of shape and
  appearance.
\newblock \emph{ArXiv}, abs/1806.06503, 2018.

\bibitem[Smidt(2020)]{smidt_2020}
Smidt, T.
\newblock Euclidean symmetry and equivariance in machine learning.
\newblock \emph{ChemRxiv}, 2020.
\newblock \doi{10.26434/chemrxiv.12935198.v1}.

\bibitem[Smith et~al.(2017)Smith, Isayev, and Roitberg]{smith2017ani}
Smith, J.~S., Isayev, O., and Roitberg, A.~E.
\newblock Ani-1: an extensible neural network potential with dft accuracy at
  force field computational cost.
\newblock \emph{Chemical science}, 8\penalty0 (4):\penalty0 3192--3203, 2017.

\bibitem[Thomas et~al.(2018)Thomas, Smidt, Kearnes, Yang, Li, Kohlhoff, and
  Riley]{thomas2018tensor}
Thomas, N., Smidt, T., Kearnes, S., Yang, L., Li, L., Kohlhoff, K., and Riley,
  P.
\newblock Tensor field networks: Rotation- and translation-equivariant neural
  networks for 3d point clouds, 2018.

\bibitem[Vaswani et~al.(2017)Vaswani, Shazeer, Parmar, Uszkoreit, Jones, Gomez,
  Kaiser, and Polosukhin]{vaswani2017attention}
Vaswani, A., Shazeer, N., Parmar, N., Uszkoreit, J., Jones, L., Gomez, A.~N.,
  Kaiser, {\L}., and Polosukhin, I.
\newblock Attention is all you need.
\newblock In \emph{Advances in neural information processing systems}, pp.\
  5998--6008, 2017.

\bibitem[Wang et~al.(2012)Wang, Asafi, van Kaick, Zhang, Cohen-Or, and
  Chen]{Wang2012ActiveCO}
Wang, Y., Asafi, S., van Kaick, O.~M., Zhang, H., Cohen-Or, D., and Chen, B.
\newblock Active co-analysis of a set of shapes.
\newblock \emph{ACM Transactions on Graphics (TOG)}, 31:\penalty0 1 -- 10,
  2012.

\bibitem[Weiler \& Cesa(2019{\natexlab{a}})Weiler and
  Cesa]{NEURIPS2019_45d6637b}
Weiler, M. and Cesa, G.
\newblock General e(2)-equivariant steerable cnns.
\newblock In Wallach, H., Larochelle, H., Beygelzimer, A., d\textquotesingle
  Alch\'{e}-Buc, F., Fox, E., and Garnett, R. (eds.), \emph{Advances in Neural
  Information Processing Systems}, volume~32. Curran Associates, Inc.,
  2019{\natexlab{a}}.
\newblock URL
  \url{https://proceedings.neurips.cc/paper/2019/file/45d6637b718d0f24a237069fe41b0db4-Paper.pdf}.

\bibitem[Weiler \& Cesa(2019{\natexlab{b}})Weiler and Cesa]{weiler2019general}
Weiler, M. and Cesa, G.
\newblock General $ e (2) $-equivariant steerable cnns.
\newblock \emph{arXiv preprint arXiv:1911.08251}, 2019{\natexlab{b}}.

\bibitem[Weiler et~al.(2018{\natexlab{a}})Weiler, Geiger, Welling, Boomsma, and
  Cohen]{Weiler20183DSC}
Weiler, M., Geiger, M., Welling, M., Boomsma, W., and Cohen, T.
\newblock 3d steerable cnns: Learning rotationally equivariant features in
  volumetric data.
\newblock In \emph{NeurIPS}, 2018{\natexlab{a}}.

\bibitem[Weiler et~al.(2018{\natexlab{b}})Weiler, Geiger, Welling, Boomsma, and
  Cohen]{weiler20183d}
Weiler, M., Geiger, M., Welling, M., Boomsma, W., and Cohen, T.
\newblock 3d steerable cnns: Learning rotationally equivariant features in
  volumetric data, 2018{\natexlab{b}}.

\bibitem[Weiler et~al.(2018{\natexlab{c}})Weiler, Hamprecht, and
  Storath]{weiler2018learning}
Weiler, M., Hamprecht, F.~A., and Storath, M.
\newblock Learning steerable filters for rotation equivariant cnns.
\newblock In \emph{Proceedings of the IEEE Conference on Computer Vision and
  Pattern Recognition}, pp.\  849--858, 2018{\natexlab{c}}.

\bibitem[Winter et~al.(2021)Winter, No{\'e}, and
  Clevert]{winter2021permutation}
Winter, R., No{\'e}, F., and Clevert, D.-A.
\newblock Permutation-invariant variational autoencoder for graph-level
  representation learning.
\newblock \emph{arXiv preprint arXiv:2104.09856}, 2021.

\bibitem[Worrall et~al.(2017)Worrall, Garbin, Turmukhambetov, and
  Brostow]{worrall2017harmonic}
Worrall, D.~E., Garbin, S.~J., Turmukhambetov, D., and Brostow, G.~J.
\newblock Harmonic networks: Deep translation and rotation equivariance.
\newblock In \emph{Proceedings of the IEEE Conference on Computer Vision and
  Pattern Recognition}, pp.\  5028--5037, 2017.

\bibitem[Xu et~al.(2021)Xu, Kim, Rainforth, and Teh]{xu2021group}
Xu, J., Kim, H., Rainforth, T., and Teh, Y.~W.
\newblock Group equivariant subsampling, 2021.

\bibitem[Zaheer et~al.(2017)Zaheer, Kottur, Ravanbakhsh, Poczos, Salakhutdinov,
  and Smola]{zaheer2017deep}
Zaheer, M., Kottur, S., Ravanbakhsh, S., Poczos, B., Salakhutdinov, R., and
  Smola, A.
\newblock Deep sets.
\newblock \emph{arXiv preprint arXiv:1703.06114}, 2017.

\end{thebibliography}
\bibliographystyle{icml2022}

 \newpage
 \section*{Checklist}

 \begin{enumerate}

 \item For all authors...
 \begin{enumerate}
   \item Do the main claims made in the abstract and introduction accurately reflect the paper's contributions and scope?
     \answerYes{}
   \item Did you describe the limitations of your work?
     \answerYes{}
   \item Did you discuss any potential negative societal impacts of your work?
     \answerNA{}
   \item Have you read the ethics review guidelines and ensured that your paper conforms to them?
     \answerYes{}
 \end{enumerate}

 \item If you are including theoretical results...
 \begin{enumerate}
   \item Did you state the full set of assumptions of all theoretical results?
     \answerYes{}
         \item Did you include complete proofs of all theoretical results?
     \answerYes{}
 \end{enumerate}

 \item If you ran experiments...
 \begin{enumerate}
   \item Did you include the code, data, and instructions needed to reproduce the main experimental results (either in the supplemental material or as a URL)?
     \answerYes{}
   \item Did you specify all the training details (e.g., data splits, hyperparameters, ow they were chosen)?
     \answerYes{}
         \item Did you report error bars (e.g., with respect to the random seed after unning experiments multiple times)?
     \answerYes{}
         \item Did you include the total amount of compute and the type of resources used (e.g., type of GPUs, internal cluster, or cloud provider)?
     \answerYes{}
 \end{enumerate}

 \item If you are using existing assets (e.g., code, data, models) or curating/releasing new assets...
 \begin{enumerate}
   \item If your work uses existing assets, did you cite the creators?
     \answerYes{}
   \item Did you mention the license of the assets?
     \answerNA{}
   \item Did you include any new assets either in the supplemental material or as a URL?
     \answerNA{}
   \item Did you discuss whether and how consent was obtained from people whose data you're using/curating?
     \answerNA{}
   \item Did you discuss whether the data you are using/curating contains personally identifiable information or offensive content?
     \answerNA{}
 \end{enumerate}

 \item If you used crowdsourcing or conducted research with human subjects...
 \begin{enumerate}
   \item Did you include the full text of instructions given to participants and screenshots, if applicable?
     \answerNA{}
   \item Did you describe any potential participant risks, with links to Institutional review Board (IRB) approvals, if applicable?
     \answerNA{}
   \item Did you include the estimated hourly wage paid to participants and the total mount spent on participant compensation?
     \answerNA{}
 \end{enumerate}

 \end{enumerate}


\clearpage
\newpage
\appendix
\onecolumn

\section*{Appendix}
\section{Proofs}
\label{app:proofs}
\begin{proposition}
\label{prop:psiequivapp}
Any suitable group function $\psi: X \rightarrow G$ is $G$-equivariant at a point $x\in X$ up the stabilizer $G_x$, i.e., 
$\psi(\rho_X(g)x) \subseteq g\cdot \psi(x) G_x$.
\end{proposition}
\emph{Proof}: 
As the relation (see Property 2.2)
\begin{align}
\label{eq:definingpsi}
\rho_X(\psi(x))\delta(\eta(x)) = x~
\end{align}
must hold for any $x\in X$, it must hold for any point $x'= \rho_X(g)x$ in the orbit of $x$, which then reads
\begin{align}
\label{eq:equivar1}
    x' &=\rho_X(\psi(x'))\delta(\eta(x')) \nonumber\\
    &= \rho_X(\psi(\rho_X(g)x))\delta(\eta(x))~,
\end{align}
where we used the invariance of $\eta$. On the other hand, applying $\rho_X(g)$ to both sides of \eqref{eq:definingpsi} we have
\begin{align}
\label{eq:equivar2}
    \rho_X(g)x &= \rho_X(g)\rho_X(\psi(x))\delta(\eta(x)) \nonumber\\
    &= \rho_X(g\psi(x))\delta(\eta(x))~,
    \end{align}
since $\eta(\rho_X(g')x) = \eta(x)$ and $\rho_X(g_1)\rho_X(g_2) = \rho_X(g_1g_2)$. 
Combining \eqref{eq:equivar1} and \eqref{eq:equivar2} it follows 
that
\begin{align}
\rho_X(\psi(x)^{-1}\cdot g^{-1} \cdot \psi(\rho_X(g)x)) \delta(\eta(x)) = \delta(\eta(x))~,
\end{align}
that is, $\psi(x)^{-1}\cdot g^{-1} \cdot \psi(\rho_X(g)x)) \in G_{ \delta(\eta(x))}$. Now, since $x$ and $\delta(\eta(x))$ by assumption belong to the same orbit of $G$, it follows that they have isomorphic stabilizers, $G_{ \delta(\eta(x))} \simeq G_x$. Thus, we have shown that 
$\psi(\rho_X(g)x)) = g \cdot \psi(x) \cdot g'$, where $g' \in G_x$, which proves our claim. 
\qedsymbol
\begin{proposition}
\label{prop:surjapp}
The image of any suitable group function $\psi:X\rightarrow G$ is surjective into $\frac{G}{G_X}$, where $G_X$ is the stabilizers of all the
points of $X$.
\end{proposition}
\emph{Proof}: 
Let $x\in X$ be such that $x = \delta(\eta(x))$, that is, $\psi(x)=G_x$, the stabilizer of $x$. Note that each orbit cointains at least one such element. For any element $g\in G$ we have that, using Proposition \ref{prop:psiequivapp}, $\psi(\rho_X(g)x) = g\cdot \psi(x)\cdot \tilde{g}$, where $\tilde{g}\in G_x$. Since $\psi(x)\cdot\tilde{g}\in G_x$ as well, it then follows that the image of $\psi$ is $G$ up to an action by an element of the stabilizer $G_x$. Applying the above reasoning to every points $x\in X$, we have that 
$\Imag(\psi) = \cup_{x\in X}\frac{G}{G_x} = \frac{G}{\cap_{x\in X} G_x}$, where $\cap_{x\in X} G_x = G_X= \{g\in G | \rho_X(g)x = x, \ \forall x\in X \}$, proving our claim.\qedsymbol

\begin{lemma}
\label{lem:isomapp}
Any suitable group function $\psi$ is an isomorphism $O_x \simeq G/G_x$ for any $x\in X$, 
where $O_x\subset X$
is the orbit of $x$ with respect to $G$ in $X$.
\end{lemma}
\emph{Proof}:
Surjectivity follows directly from Proposition \ref{prop:surjapp}. 
To show injectivity, consider
$x, x'\in O_x$ such that $\psi(x') = \psi(x)\cdot \tilde{g}$, where $\tilde{g}\in G_x$.
From Proposition \ref{prop:psiequiv} it follows that $x'=x$, which proves the claim. \qedsymbol




\begin{proposition}
\label{prop:muequivapp}
Let $\psi = \xi \circ \mu$ be a suitable group function and let $\mu: X \rightarrow Y$ be $G$-equivariant. Then, $G_x = G_{\mu(x)}$ for all $x\in X$.
\end{proposition}
\emph{Proof}:
Let $g\in G_x$, that is, $\rho_X(g)x = x$. Applying $\mu$ to both sides of this equation we obtain $\mu(x) = \mu(\rho_X(g)x) = \rho_Y(g)\mu(x)$, where we used the $G$-equivariance of $\mu$.
Hence, $G_x \subseteq G_{\mu(x)}$. To prove the opposite inclusion, let $g\in G_{\mu(x)}$
but $g\notin G_x$, and let $x'=\rho_X(g)x$. Now, $\mu(x') = \rho_Y(g)\mu(x) = \mu(x)$, thus $\mu$, and therefore $\psi = \xi\circ \mu$, maps the distinct element $x, x'$
to the same group element $\psi(x) = \psi(x')$, in contradiction with Proposition \ref{prop:psiequiv}. \qedsymbol


\begin{proposition}
\label{prop:xiequivapp}
Given $y, y_0$, the element $g$ such that $y \equiv \rho_Y(g) y_0$ is unique up to the stabilizer $G_{y_0}$.
\end{proposition}
\emph{Proof}:
Suppose that there exist $g_1,g_2\in G$ such that $\rho_Y(g_1) y_0 = \rho_Y(g_2) y_0$, then 
$\rho_Y(g_2^{-1}g_1)y_0 = y_0$, which implies $g_2^{-1} g_1 \in G_{y_0}$. \qedsymbol

\begin{proposition}
Let $\psi = \xi \circ \mu$ where $\mu$ and $\xi$ are as described above. Then, 
$\psi$ is a suitable group function.
\end{proposition}
\emph{Proof}:
We show that our construction describes an isomorphism $O_x \simeq G/G_x$ for all $x\in X$.
Given $x\in X$ and $g\in G$, Propositions \ref{prop:muequivapp} and 
\ref{prop:xiequivapp} imply
\begin{align}
\label{eq:eqpropconstr}
\xi\left(\mu(\rho_X(g)x)\right) = \xi(\rho_Y(g) \mu(x))\subseteq g\cdot\xi(\mu(g)) G_x~,
\end{align}
that is, $\psi$ possesses the $G$-equivariant property as required in Proposition
\ref{prop:psiequiv}, which in turns imply injectivity, as in Lemma \ref{lem:isomapp}.
Surjectivity follows from the same argument as in Proposition \ref{prop:surjapp}, since
the proof only relies on the equivariant properties of $\psi$, which we showed in 
\eqref{eq:eqpropconstr}. \qedsymbol

\section{Model architecture Rotated MNIST}
\label{app:mnist}
We follow \cite{weiler2019general} and use steerable CNNs to parameterize functions $\eta$ and $\mu$. In contrast to classical CNNs, CNNs with O(2)-steerable kernels transform feature fields respecting the transformation law under actions of $O(2)$. We can define scalar fields $s: \mathbb{R}^2 \rightarrow \mathbb{R}$ and vector fields $v: \mathbb{R}^2 \rightarrow \mathbb{R}^2$ that transform under group actions (rotations) the following:
\begin{equation}
    s(x) \mapsto s(g^{-1}x) \qquad v(x) \mapsto g\cdot v(g^{-1}x) \qquad \forall g \in O(2)~.
\end{equation}
Thus, scalar values are moved from one point on the plane $\mathbb{R}^2$ to another but are not changed, while vectors are moved and changed (rotated) equivalently. Hence, we can utilize steerable CNNs to encode samples in $\mathbb{R}^2$ in $O(2)$-invariant scalar features and $O(2)$-equivariant vector features. We can use the scalar features $s$ as $\mathcal{G}$-invariant representation $z\in Z$ and following Section 3 (Orthogonal group $\SO(2)$) utilizing a single vector features $v$ to construct the rotation matrix $R$ as:
\begin{equation}
    R = \begin{bmatrix}
        \bar{v}_x & -\bar{v}_y\\
        \bar{v}_y & \bar{v}_x
\end{bmatrix}, \qquad \bar{v} = \frac{v}{\|v\|} .
\end{equation}
In our experiments we used seven layers of steerable CNNs as implemented by \cite{weiler2019general}. We did not use pooling layers, as we found them to break rotation equivariance and only averaged over the two spatial dimensions after the final layer to extract the final invariant embedding and equivariant vector. In each layer we used 32 hidden scalar and 32 hidden vector fields. In the final layer we used 32 scalar fields (32 dimensional invariant embedding) and one vector feature field.

The Decoding function $\delta:Z\rightarrow\mathbb{R}^2$ can be parameterized by a regular CNN. In our experiments we used six layers of regular CNNs with 32 hidden channels, interleaved with bilinear upsampling layers starting from the embedding expanded to a $2\times 2 \times 32 $ tensor.

Training was done on one NVIDIA Tesla V100 GPU in approximately 6 hours.

\section{Model architecture Set of Digits}
\label{app:digits}
We can rewrite the equation $P_{\sigma}(1,2,\dots, n) = (\sigma(1), \sigma(2), \dots, \sigma(n))~,$ in vector form by representing set elements by standard $n \times 1$ column vectors $\mathbf{e}_i$ (one-hot encoding) and $\sigma$ by a permutation matrix $P_\sigma$ whose (i,j) entry is $1$ if $i=\sigma(j)$ and $0$ otherwise, then:
\begin{equation}
    P_\sigma \mathbf{e}_i = \mathbf{e}_{\sigma(i)}
\end{equation}
Hence, encoding function $\eta$ should encode a set of elements in a permutation invariant way and $\psi$ should map a set $M$ to a permutation matrix $P_\sigma$:
\begin{equation}\label{eq:predict-permutation-matrix}
    \psi:M\rightarrow P_\sigma
\end{equation}
We follow \cite{zaheer2017deep} and parameterize $\eta$ by a neural network $\gamma$ that is applied element-wise on the set followed by an invariant aggregation function $\Sigma$ (e.g. sum or average) and a second neural network $\beta$:
\begin{equation}
    \eta(X) = \beta(\Sigma_{x\in X}\gamma(x))~.
\end{equation}
In our experiments we parameterized $\gamma$ and $\beta$ with regular feed-forward neural networks with three layers respectively, also using ReLU activations and Batchnorm.

The output of function $\gamma$ is equivariant and can also be used to construct $\psi$. We follow \cite{winter2021permutation} and define a function $s:\mathbb{R}^d\rightarrow \mathbb{R}$ mapping the output of $\gamma$ for every set element to a scalar value. By sorting the resulting scalars, we construct the permutation matrix $P_\sigma$ with entries $p_{ij}$ that would sort the set of elements with respect to the output of $s$:
\begin{equation}
    p_{ij} =
    \begin{cases}
        1, &\text{if $j=$ argsort$(s)_i$}\\
        0, &\text{else}
    \end{cases}
\end{equation}
As the argsort operation is not differentiable, we utilizes a continuous relaxation of the argsort operator proposed in \citep{prillo2020softsort, grover2019stochastic}:
\begin{equation}
\label{softsort}
    \mathbf{P} \approx \hat{\mathbf{P}} = \text{softmax}(\frac{-d(\text{sort}(s)1^\top, 1 s^\top)}{\tau}),
\end{equation}
where the softmax operator is applied row-wise, $d(x, y)$ is the $L_1$-norm and $\tau\in\mathbb{R}_+$ a temperature-parameter. \\
Decoding function $\delta$ can be parameterized by a neural network that maps the permutation-invariant set representation back to either the whole set or single set elements. In the letter case, where the same function is used to map the same set representation to the different elements, additional fixed position embeddings can be fed into the function to decode individual elements for each position/index. For the reported results we choose this approach, using one-hot vectors as position embeddings and a 4-layer feed-forward neural network.

Training was done on one NVIDIA Tesla V100 GPU in approximately 1 hours. 

\section{Model architecture Point Cloud - Tetris 3D \& QM9}
\label{app:pointcloud}
We implement a graph neural network (GNN) that transform equivariantly under rotations and translations in 3D space, respecting the invariance and equivariance constraints mentioned in Eq. \eqref{eq:rotason} and \eqref{eq:translation-vector-se(n)} for $n=3$.

Assume we have a point cloud of $N$ particles each located at a certain position $x_i \in \R^{3}$ in Cartesian space. 
Now given some arbitrary ordering $\sigma (\cdot)$ for the points, we can store the positional coordinates in the matrix $P=[x_1, ..., x_N] \in \R^{N\times 3}$. Standard Graph Neural Networks (GNNs) perform message passing \cite{gilmer2017neural} on a local neighbourhood for each node. Since we deal with a point cloud, common choice is to construct neighbourhoods through a distance cutoff $c>0$. 
The edges of our graph are specified by \textit{relative} positions 
\begin{align*} \label{eq:relative_position}
    x_{ij} = x_j - x_i \in \R^3~, 
\end{align*}
 and the neighbourhood of node $i$ is defined as 
$\mathcal{N}(i) = \{j : ~d_{ij} := ||x_{ij}|| \leq c\}$.

Now, our data (i.e., the point cloud) lives on a vector space $X$, where we want to learn an SE(3) invariant and equivariant embedding wrt. arbitrary rotations and translations in $3$D space. Let the feature for node $i$ consist of an invariant (type-0) embedding $h_i \in \R^{F_s}$, an equivariant (type-1) embedding $w_{i} \in \R^{3 \times F_v}$ that transforms equivariantly wrt. arbitrary rotation \textbf{but} is invariant to translation. Such a property can be easily obtained, when operating with relative positions.\\
Optionally, we can model another equivariant (type-1) embedding $t_i \in \R^{3}$ which transforms equivariantly wrt. translation \textbf{and} rotation.
As our model needs to learn to predict group actions in the SE(3) symmetry, we require to predict an equivariant translation vector ($b \in T_3$), as well as a rotation matrix ($A \in \text{SO}(3)$), where we will dedicate the $t$ vector to the translation and the $w$ vector(s) to the rotation matrix.\\
As point clouds might not have initial features, we initialize the SE(3)-invariant embeddings as one-hot encoding $h_i = \mathbf{e}_i$ for each node $i=1, \dots, N$. The (vector) embedding dedicated for predicting the rotation matrix is initialized as zero-tensor for each particle, i.e.,  $w_i = \mathbf{0}$ and the translation vector is initialized as the absolute positional coordinate, i.e. to, $t_i = x_i$. \\

We implement following edge function $\phi_e: \R^{2F_s + 1} \mapsto \R^{F_s + 2F_v + k}$ with
\begin{equation}\label{eq: message}
    {m}_{ij} = \phi_{e}(h_i, h_j, d_{ij}) = W_e [h_i, h_j, d_{ij}] + b_e,
\end{equation}
and set $k=1$ if the GNN should model the translation and $k=0$ else.
Notice that the message ${m}_{ij}$ in Eq. \eqref{eq: message} only depends on SE(3) invariant embeddings. 
Now, (assuming $k=1$) we further split the message tensor into 4 tensors,
\begin{align*}
    m_{ij} = [m_{h,ij}, m_{w_0, ij}, m_{w_1, ij}, m_{t, ij}]~,
\end{align*}
which we require to compute the aggregated messages for the SE(3) invariant and equivariant node embeddings.\\
We include a row-wise transform $\phi_s: \R^{F_s} \mapsto \R^{F_s}$ for the invariant embeddings using a linear layer:
\begin{equation}
    \tilde{h}_i = W_s h_i + b_s~, 
\end{equation}
The aggregated messages for invariant (type-0) embedding $h_i$ are calculated using:
\begin{equation}
    m_{i, h} = \sum_{j\in \mathcal{N}(i)} m_{h, ij} \odot \tilde{h}_i~~\in \R^{F_s}.
\end{equation}
where $\odot$ is the (componentwise) scalar-product.\\
The aggregated equivariant features are computed using the tensor-product $\otimes$ and scalar-product $\odot$ from (invariant) type-0 representations with (equivariant) type-1 representations:
\begin{equation}\label{eq:vec_message}
    m_{i, w} = \sum_{j \in \mathcal{N}(i)} \left ( x_{ij} \otimes m_{w_0, ij} + (w_{i} \times w_{j})  \odot (\mathbf{1} \otimes m_{w_1, ij}) \right) ~~\in \R^{3 \times F_v}~,
\end{equation}
where $\mathbf{1}\in \R^3$ is the vector with $1$'s as components and $(a \times b)$ denotes the cross product between two vectors $a,b\in \R^3$.\\
The tensor in Eq. \eqref{eq:vec_message} is equivariant to arbitary rotations and invariant to translations. It is easy to prove the translation invariance, as any translation $t^* \in T_3$ acting on points $x_i, x_j$ does not change the relative position $x_{ij} = (x_j + t^*) - (x_i + t^*) = x_j - x_i$.\\
To prove the rotation equivariance, we first observe that given any rotation matrix $A \in \textsc{SO}(3)$ acting on the provided data, as a consequence relative positions rotate accordingly, since
\begin{align*}
    Ax_j - Ax_i = A(x_j - x_i) = Ax_{ji} \in \R^{3}.
\end{align*}
The tensor product $\otimes$ between two vectors $u\in \R^3$ and $v\in \R^{F_s}$, commonly also referred to as \textit{outer product} is defined as
\begin{align*}
    u \otimes v = u v^\top \in \R^{3 \times F_s}~,
\end{align*}
and returns a matrix given two vectors. For the case that a group representation of SO(3), i.e. a rotation matrix $R$, acts on $u$, it is obvious to see with the associativity property
\begin{align*}
    (Au) \otimes v = (Au)v^\top = Auv^\top = A(uv^\top) = A(u\otimes v)~ = Au \otimes v.
\end{align*}\\
The cross product $(w_i \times w_j) \in \R^{3\times F_v}$ used in equation \eqref{eq:vec_message} between type-1 features $w_i$ and $w_j$ is applied separately on the last axis. The cross product has the algebraic property of rotation invariance, i.e. given a rotation matrix $A$ acting on two 3-dimensional vectors $a,b\in\R^3$ the following holds:
\begin{equation}
    (Aa) \times (Ab) = A(a\times b)~.
\end{equation}\\
Now, notice that the quantities that "transform as a vector" which we call type-1 embeddings are in $S=\{x_{ij}, w_i, t_i\}_{i,j=1}^N$.\\
Given a rotation matrix $A$ acting on elements of $S$, we can see that the result in \eqref{eq:vec_message}
\begin{align*}
     & \sum_{j \in \mathcal{N}(i)} \left ( Ax_{ij} \otimes m_{w_0, ij} + (Aw_{i}) \times (Aw_{j}) \odot (\mathbf{1} \otimes m_{w_1, ij}) \right) \\
      & = \sum_{j \in \mathcal{N}(i)} \left ( Ax_{ij} \otimes m_{w_0, ij} + A(w_{i} \times w_{j}) \odot (\mathbf{1} \otimes m_{w_1, ij}) \right) \nonumber \\
     & = A \sum_{j \in \mathcal{N}(i)}  \left ( x_{ij} \otimes m_{w_0, ij} + (w_i \times w_{j}) \odot (\mathbf{1} \otimes m_{w_1, ij}) \right)\nonumber \\
     & = A m_{i, w}
\end{align*}
is rotationally equivariant.\\
We update the hidden embedding with a residual connection
\begin{align}
    h_i &\xleftarrow{} h_i + m_{i, h}~, \nonumber \\
    w_i &\xleftarrow{} w_i + m_{i, w}~, 
\end{align}
and use a Gated-Equivariant layer with equivariant non-linearities as proposed in the PaiNN architecture \cite{schuett2021equivariant} to enable an information flow between type-0 and type-1 embeddings.\\
The type-1 embedding for the translation vector is updated in a residual fashion
\begin{align}\label{eq:translation-vec}
    t_i &\xleftarrow[]{} t_i + \sum_{j\in \mathcal{N}(i)} x_{ij} \otimes m_{t, ij} \nonumber \\
    &= t_i + \sum_{j\in \mathcal{N}(i)} m_{t, ij} \odot  x_{ij}, ~
\end{align}
where we can replace the tensor-product with a scalar-product, as $m_{t,ij}\in \R$. The result in Eq. \eqref{eq:translation-vec} is translation and rotation equivariant as the first summand $t_i$ is rotation and translation equivariant, while the second summand is only rotation equivariant since we utilize relative positions.

For the SE(3) Tetris experiment, the encoding function $\eta: X \mapsto Z$ is a 5-layer GNN encoder with $F=F_s=F_v=32$ scalar- and vector channels and implements the translation vector, i.e. $k=1$.\\
The encoding network $\eta$ outputs four quantities: two SE(3) invariant node embedding matrices $\widetilde{H}, M \in \R^{N \times F}$, one SO(3) equivariant order-3 tensor $\widetilde{W}\in \R^{N\times 3\times F}$ as well as another SE(3) equivariant matrix $T\in \R^{N\times 3}.$ \\
We use two linear layers\footnote{The transformation is always applied on the last (feature) axis.} to obtain the SE(3) invariant embedding matrix $H \in \R^{N\times 2}$ as well as the SO(3) equivariant embedding tensor $W\in \R^{N\times 3\times 2}$. Notice that the linear layer returning the $W$ tensor can be regarded as the function $\psi_\text{rot}$ that aims to predict the group action in the SO(3) symmetry, while we use the identity map for the translation vector, i.e. $\psi_\text{transl} = T$.\\
As point clouds can be regarded as sets, we obtain an permutation invariant embedding by averaging over the first dimension of the $\{H, W, T\}$ tensors,
\begin{align}
    &h = \frac{1}{N} \sum_{i=1}^N H_i ~\in \R^2~, \\
    &t = \frac{1}{N} \sum_{i=1}^N T_i ~\in \R^3~,\\
    &w = \frac{1}{N} \sum_{i=1}^N W_i ~\in \R^{3\times 2}~,
\end{align}
while we use the $M$ matrix to predict the permutation matrix $P_\sigma$ with the $\psi_{\text{perm}}$ function, in similar fashion as described in Eq. \eqref{eq:predict-permutation-matrix}. To construct the rotation matrix $R$ out of 2 vectors in $\R^3$ as described in Section \ref{examples}, we utilize the SO(3) equivariant embedding $w$.\\
The decoding network $\delta: Z \mapsto X$ is similar to the encoder a $5$-layer SE($3$)-equivariant GNN but does not model the translation vector, i.e. $k=0$. The decoder $\delta$ maps the SE($3$) as well as S($N$)-invariant embedding $h$ back to a reconstructed point cloud $\hat{P}\in \R^{N\times 3}$. At the start of decoding, we utilize a linear layer to map the $G-$invariant embedding $h\in \R^2$ to a higher-dimension, i.e.
\begin{equation}
    \tilde{h} = W_0h + b_0 ~\in \R^{F_s},
\end{equation}
Next, to ``break'' the symmetry and provide the nodes with initial type-0 features, we utilize fixed (deterministic) positional encodings as suggested by \cite{winter2021permutation} for each node $i=1, \dots, N$ to be summed with $\tilde{h}$. Notice that this addition enables us to obtain distinct initial type-0 embeddings $\{\hat{h}_i\}_{i=1}^{N}$. \\ For the start positions, we implement a trainable parameter matrix $P_\theta$ of shape $(N \times 3)$ for the decoder.\\
Now, given an initial node embedding $\hat{H} \in \R^{N\times F_s}$, we apply the S($N$) group action, by multiplying the predicted permutation matrix $P_\sigma$ with $\hat{H}$ from the left to obtain the canonical ordering as
\begin{align}
    &\hat{H}_\sigma = P_\sigma \hat{H}~.
\end{align}
To retrieve the correct orientation required for the pairwise-reconstruction loss, we multiply the constructed rotation matrix $R$ with the initial start position matrix ${P}_\theta$
\begin{equation}
    \hat{P}_{r} = {P}_{\theta} R^\top.
\end{equation}
With such construction, we can feed the two tensors to the decoder network $\delta$ to obtain the reconstructed point cloud as
\begin{equation}
    \hat{P}_\text{recon} = \delta(\hat{H}_\sigma, \hat{P}_{r}) + t ~, 
\end{equation}
where $t \in \R^3$ is the predicted translation vector from the encoder network, added row-wise for each node position.

\section{Further Experiments}

\subsection{Rotated MNIST}

\begin{figure}
     \centering
     \includegraphics[width=0.99\textwidth]{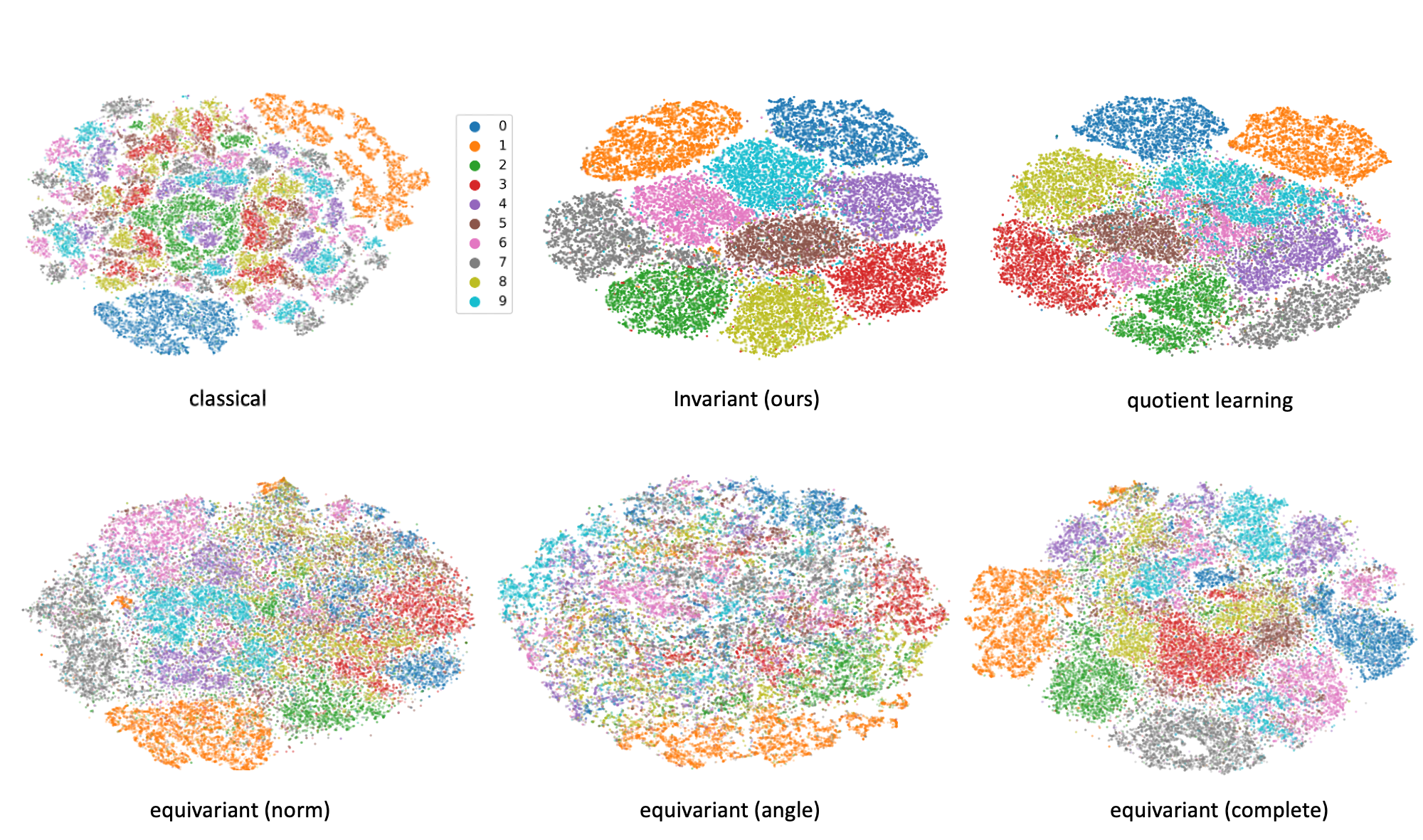}
     \caption{Top row: TSNE embedding of the encoded test dataset for a classical autoencoder,  our proposed SO(2) invariant autoencoder, and for the quotient autoencoder of \cite{mehr2018manifold}. Bottom row: Fully equivariant trained autoencoder with invariant projection after training, either by taking the norm, angles between vectors or the combination (complete).}
     \label{fig:mnist_new}
\end{figure}

We also implemented and trained the quotient autoencoder (QAE) approach proposed by \cite{mehr2018manifold} on the MNIST dataset for the group $\SO(2)$, discretized in 36 rotations with the loss
\begin{align}
    \min_{\theta \in \{10i, i=0,\dots, 35\}} \left\{ \text{MSE}(x - \rho_X(g(\theta))y) \right\}~,
\end{align}
where $x$ is a MNIST sample and $y$ is the reconstructed sample.
We evaluated the resulting embeddings on the rotated MNIST test set (in such a way that the evaluation is the same as for our model). In Figure \ref{fig:mnist_new} we plot TSNE embeddings for this approach, and we can observe that the embedding space shows a clearer structure, in comparison with the classical model. However, in comparison, our approach results in a better clustering of the different digits classes. That shows that the discretization step, while it helps in structuring the embedding space in ``signal clusters'', still does not capture the full continuous nature of the group. To further quantitatively compare the three methods (ours, QAE and classical AE), we evaluated the reconstruction loss as well as the (digit class) classification accuracy of a KNN classifier trained on 1000 embeddings of each method.
We present in the table below the results for the reconstruction loss and for the classification accuracy of a KNN classifier trained on the AE embeddings. To obtain a fair comparison, we kept the architecture and the training hyperparameters exactly identical for all the strategies. We note that our strategy outperforms both the classical AE as well as the strategy of QAE in both tasks.

In an additional experiment, we trained a fully equivariant AE (that is, the embedding itself is fully equivariant, i.e. multiple 2-dimensional vectors)
on MNIST with $G=\SO(2)$, followed by an invariant pooling afterwards (after the training) to extract the invariant part. 
Specifically, we have trained KNN classifiers on (a) the invariant embedding corresponding to the norm of the 2-dimensional vectors forming the bottleneck representation, (b) the angles  between the first and all other vectors and on (c) the full invariant embedding we obtained by combining the the norms and angles. We choose the number of vectors in the bottleneck in such a way that the dimensionality of the full invariant representation coincides with the one of our model. We visualized the resulting TSNE embeddings in Figure \ref{fig:mnist_new}
and show the downstream performance of the KNN classifiers in Table \ref{tab:mnist}.
From the results we can see that, in comparison to the approximate invariant (QAE) and our invariant trained model, the invariant projected equivariant representations perform inferior. Although we extract a complete invariant representation (which performs better than a subset of this representation like the norm or angle part), the resulting representation is apparently not as expressive and e.g. useful in a downstream classification task. This aligns well with our hypothesis, that our proposed framework poses a sensible supervisory signal to extract expressive invariant representations that are superior to invariant projections of equivariant features. 
\begin{table}
\caption{Comparison of our approach vs classical and quotient autoencoder (QAE) as well as an fully equivariant AE with invariant pooling after training.}
\label{tab:mnist}
\begin{center}
\begin{tabular}{||c | c c||} 
 \hline
 Model & Rec.~Loss & KNN Acc. \\ [0.5ex] 
 \hline\hline
 classical & 0.0170 & 0.68\\ 
 QAE & 0.0227 & 0.82\\ 
invariant~(ours) & 0.0162 & 0.90\\
equiv AE (norm)  & 0.0189 & 0.56 \\ 
equiv AE (angle)  & 0.0189 & 0.53 \\ 
equiv AE (complete)  & 0.0189 & 0.67 \\ 
 \hline
\end{tabular}
\end{center}
\end{table}

\subsection{QM9}

\begin{table}[t!]
    \centering
    \begin{tabular}{||c | c c c||}
    \hline
     Target& Fraction  & Pretrained   & From Scratch  \\
     \hline
     \multirow{ 2}{*}{$H$} & $0.05$ & $0.7529$  & $0.0970$      \\
      & $0.25$ & $0.9908$ & $0.9093$     \\
     \hline
     \multirow{ 2}{*}{$G$} & $0.05$ & $0.7703$ & $0.4758$    \\
         & $0.25$ & $0.9856$ & $0.9751$    \\
     \hline
     \multirow{ 2}{*}{$U$} & $0.05$ & $0.6083$ & $0.2574$    \\
      & $0.25$ & $0.9962$ & $0.9808$    \\
     \hline
     \multirow{ 2}{*}{$\langle R^2 \rangle$} & $0.05$ & $0.7806$ & $0.1468$    \\
      & $0.25$ & $0.9918$ & $0.8546$    \\
     \hline
     \multirow{ 2}{*}{$\mu$} & $0.05$ &  $0.8698$ &  $0.8443$   \\
     & $0.25$ &  $0.9718$ & $0.9718$    \\
     \hline
     \multirow{ 2}{*}{$\alpha$} & $0.05$ & $0.9455$  & $0.9237$     \\
     & $0.25$ &  $0.9937$ & $0.9764$    \\
     \hline
    \end{tabular}
    \caption{Generalization performance in terms of the coefficient of determination $R^2$ of models on a held-out test set of $1000$ samples. Higher $R^2$ indicates better performance.}
    \label{table:pretraining}
\end{table}

For the QM9 dataset, we use the same model components as described in the Tetris experiment, with the difference of including atom species as $SE(3)-$invariant features and setting $F_s = 256, F_v=32$ and increasing the dimensionality of the latent space to $256$.

\subsubsection{Finetuning}
We performed additional experiments on the pretrained group-invariant AE on the extended GEOM-QM9 dataset \cite{geom-axelrod} which, as opposed to the standard QM9 dataset ($\approx 130k$ samples), contains multiple conformations of small molecules. We trained the autoencoder on a reduced set of GEOM-QM9 ($\approx 641k$), containing up to $10$ conformations per molecule and utilized this pretrained encoder network to regress (invariant) energy targets, such as internal energy $U$ or enthalpy $H$ on the original QM9 dataset.

\begin{figure}[t!]
    \centering
    \includegraphics[width=\columnwidth]{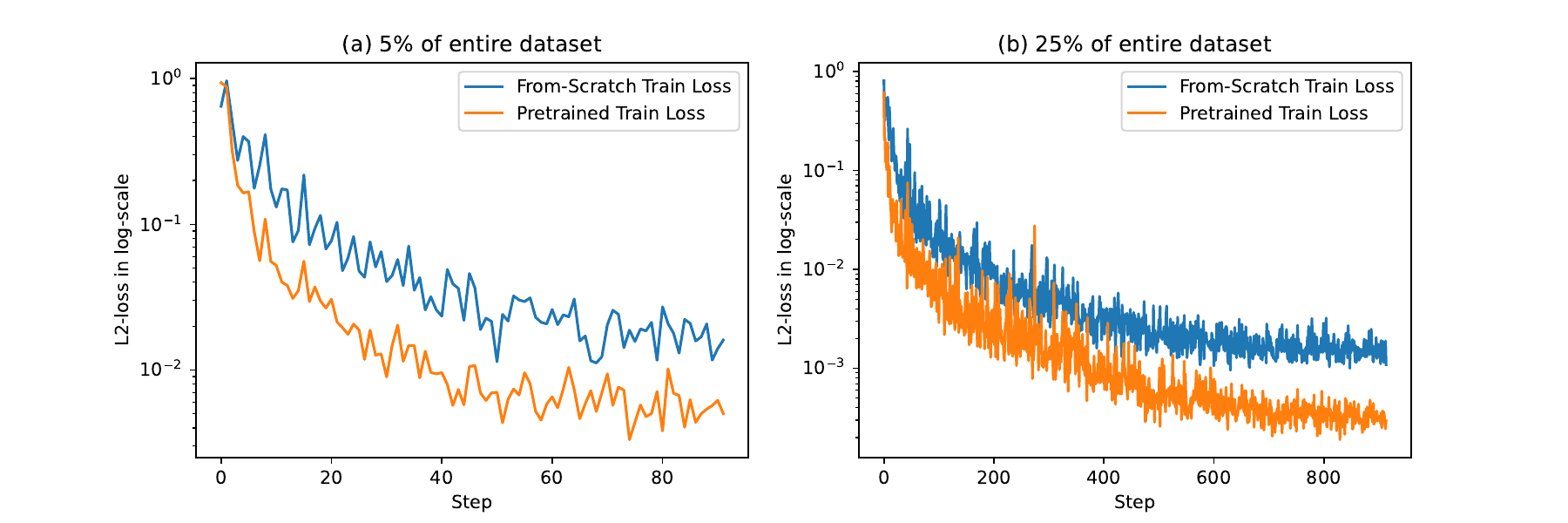}
    \caption{Learning curves of models trained on limited enthalpy $H$ targets.}
    \label{fig:learning-curves}
\end{figure}

We observed that the pretrained encoder network learns faster and achieves better generalization performance than the architectural identical network trained from scratch. In Figure \ref{fig:learning-curves} we illustrate the learning curves for the two networks on different fraction on $5\%$ and $25\%$ labelled samples from the original QM9 dataset to analzye the benefit of finetuning a pre-trained encoder network in a low-data regime, when regressing on the enthalpy $H$.
On a held-out test dataset of $1000$ samples, the pretrained encoder network achieves superior generalization performance in terms of $R^2$ with $0.7529$ vs. $0.0970$ in the $5\%$ data regime, and $0.9908$ vs. $0.9093$ in the $25\%$ data regime compared to the encoder that was trained from scratch.
In Table\ref{table:pretraining} we show additional comparisons of the pretrained network against a network that was trained from scratch for 50 epochs on the restricted dataset.

As shown in Table \ref{table:pretraining}, the pretrained encoder achieves improved generalization performance on the test dataset compared to its architectural identical model that was trained from scratch. We believe that training the group-invariant autoencoder on a larger diverse dataset of (high-quality) molecular conformations facilitates new opportunities in robust finetuning on different data-scarse datasets for molecular property prediction.

\subsubsection{Molecular Conformations: Further Examples}
\label{app:qm9examples}

\begin{figure}
    \centering
    \includegraphics[width=\columnwidth]{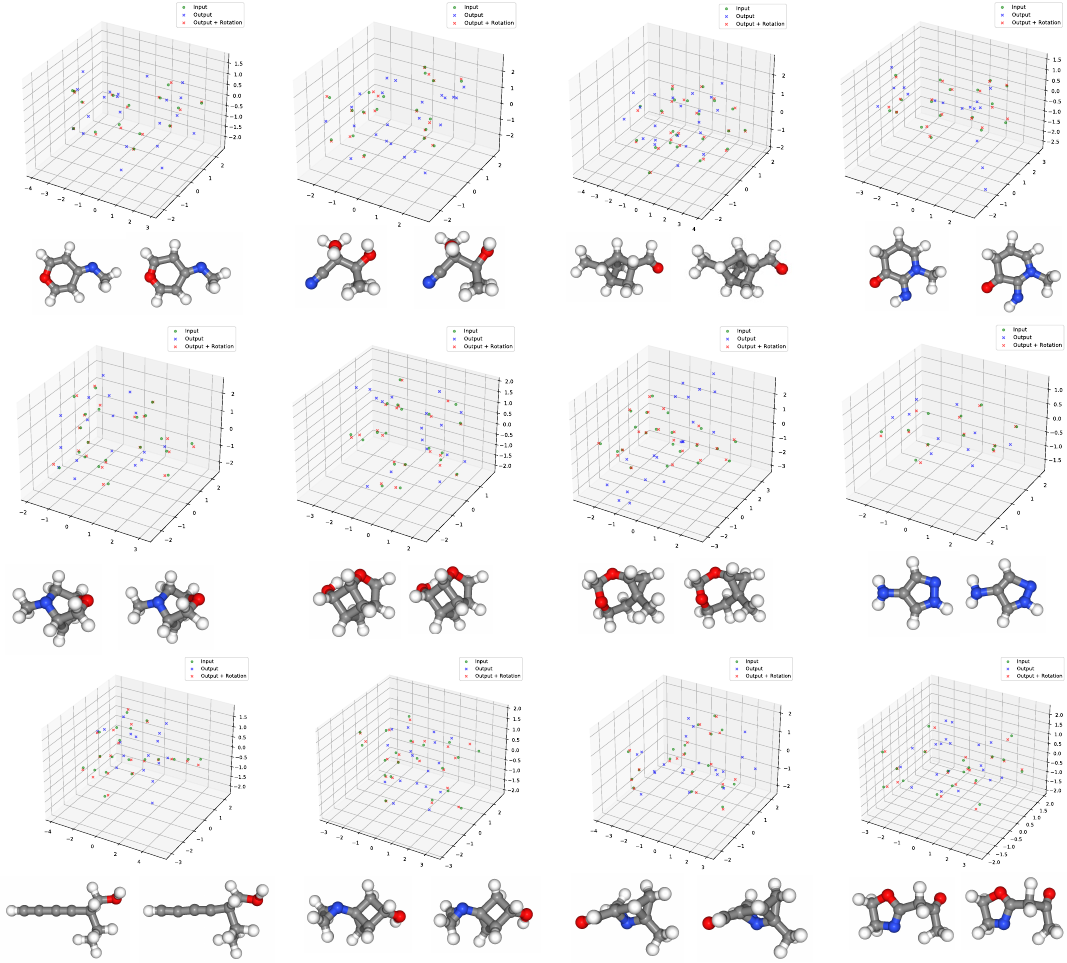}
    \caption{12 molecular conformations and their reconstructions represented as point cloud and ball-and-stick model (left true, right predicted).}
    \label{fig:additional-recons}
\end{figure}

\begin{figure}
    \centering
    \includegraphics[width=0.5\columnwidth]{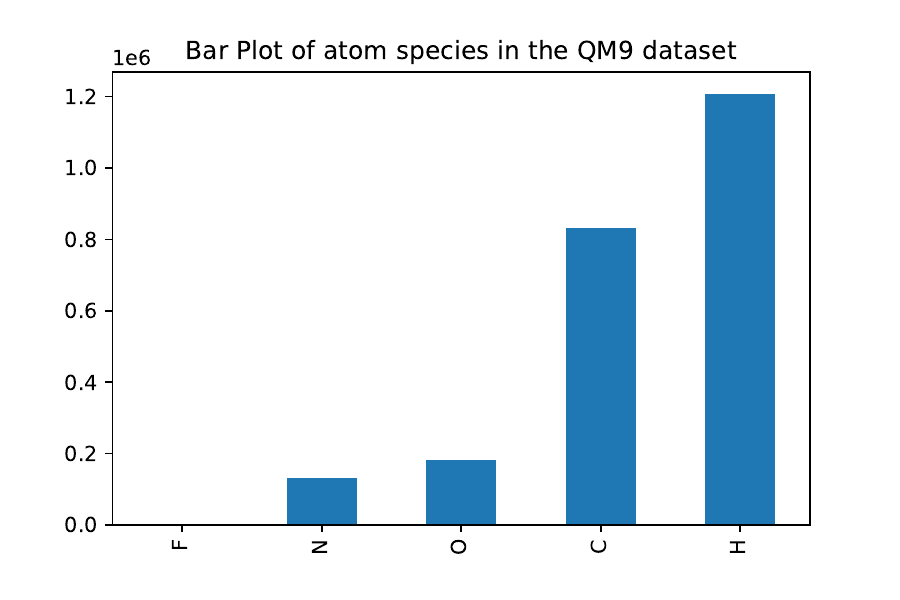}
    \caption{Atomistic species count on the QM9 dataset.}
    \label{fig:barplot}
\end{figure}

We show additional reconstructions of 12 randomly selected small molecules from the QM9 test dataset. Noticeably, our trained autoencoder is able to reconstruct molecular conformations with complex geometries as depicted in the third column (from the left). We notice that the AE is not able to perfectly reconstruct the conformation shown in the 4th column of the 2nd row. Although this molecule does not exhibit a complicated geometrical structure, its atomistic composition (of only containing nitrogen and carbon as heavy atoms) could be the reason why the encoding of the conformation is pointing into a non-densely populated region in the latent space, as nitrogen does not have a large count in the total QM9 database, see Figure \ref{fig:barplot}.

Training was done on one NVIDIA Tesla V100 GPU in approximately 1 day.

\begin{figure}
    \centering
    \includegraphics[width=1.0\columnwidth]{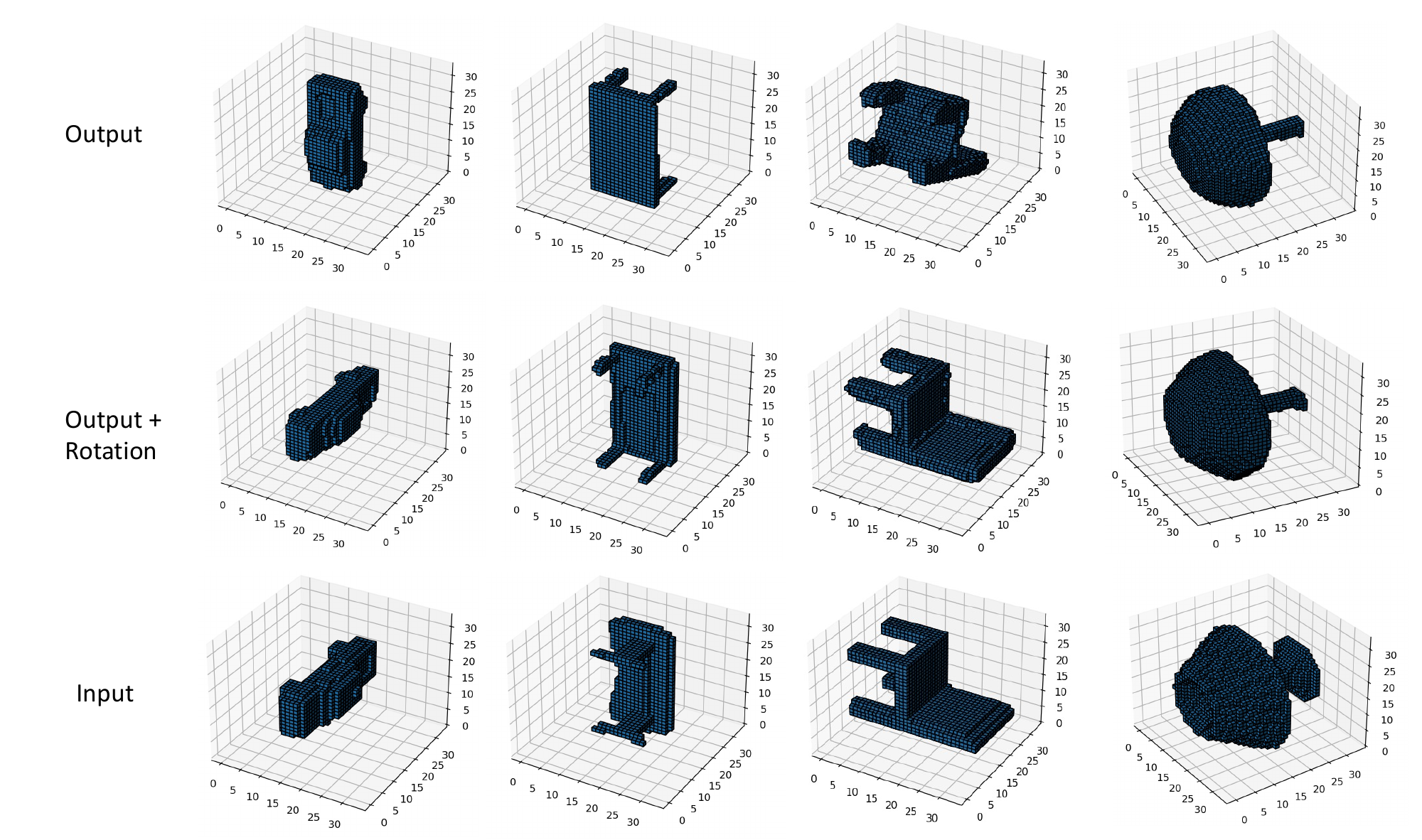}
    \caption{Example reconstructions from our proposed method trained on the voxelized ShapeNet dataset \cite{chang2015shapenet}. }
    \label{fig:shapenet_rec}
\end{figure} 

\end{document}